\documentclass{arxivtmpl}

\def\1{\bm{1}}

\DeclareMathAlphabet{\mathsfit}{\encodingdefault}{\sfdefault}{m}{sl}
\SetMathAlphabet{\mathsfit}{bold}{\encodingdefault}{\sfdefault}{bx}{n}

\tcbuselibrary{skins,breakable}
\newtcolorbox{findingbox}{enhanced,breakable,frame hidden,colback=gred!6,borderline west={3pt}{0pt}{gred},sharp corners,boxsep=0pt,left=12pt,right=10pt,top=8pt,bottom=8pt,before skip=12pt,after skip=12pt}
\newtcolorbox{notebox}{enhanced,breakable,frame hidden,colback=gblue!6,borderline west={3pt}{0pt}{gblue},sharp corners,boxsep=0pt,left=12pt,right=10pt,top=8pt,bottom=8pt,before skip=12pt,after skip=12pt}
\newtcolorbox{tipbox}{enhanced,breakable,frame hidden,colback=ggreen!6,borderline west={3pt}{0pt}{ggreen},sharp corners,boxsep=0pt,left=12pt,right=10pt,top=8pt,bottom=8pt,before skip=12pt,after skip=12pt}
\newtcolorbox{warnbox}{enhanced,breakable,frame hidden,colback=gyellow!6,borderline west={3pt}{0pt}{gyellow},sharp corners,boxsep=0pt,left=12pt,right=10pt,top=8pt,bottom=8pt,before skip=12pt,after skip=12pt}

\tcbuselibrary{listings}
\newcommand{\boxfont}{\fontsize{9}{11}\selectfont}
\newtcblisting{promptbox}[1][]{%
  enhanced, listing only, breakable, frame hidden, width=\linewidth,
  colback=gblue!6, borderline west={3pt}{0pt}{gblue},
  sharp corners, boxsep=0pt,
  left=12pt, right=10pt, top=8pt, bottom=8pt, before skip=12pt, after skip=12pt,
  before upper={\def\promptlabel{#1}\ifx\promptlabel\empty\else{\sffamily\bfseries\color{gblue}#1}\par\smallskip\fi},   
  listing options={basicstyle=\ttfamily\boxfont, escapeinside={(*}{*)}},
}

\newtcolorbox{transcriptbox}{%
  enhanced, breakable, frame hidden,
  colback=ggreen!6, borderline west={3pt}{0pt}{ggreen},
  sharp corners, boxsep=0pt, fontupper=\ttfamily\boxfont, halign upper=flush left,   
  left=12pt, right=10pt, top=6pt, bottom=8pt, before skip=12pt, after skip=12pt,
}

\colorlet{codebg}{gyellow!6}
\setminted{fontsize=\boxfont, breaklines, autogobble, tabsize=4, style=vs}
\newminted[codebox]{python}{bgcolor=codebg, frame=leftline, framerule=2.5pt, rulecolor=gyellow, framesep=2.5mm}

\hypersetup{
    colorlinks=true,
    linkcolor=linkcol,
    citecolor=linkcol,
    urlcolor=linkcol,
    pdfborderstyle={},
    pdfborder={0 0 0}
}
\renewcommand\Affilfont{\rmfamily\mdseries\fontsize{11.5}{14}\selectfont}

\colorlet{perfhue}{gblue}
\colorlet{perf0}{white}
\colorlet{perf10}{perfhue!12}
\colorlet{perf20}{perfhue!22}
\colorlet{perf30}{perfhue!33}
\colorlet{perf40}{perfhue!44}
\colorlet{perf50}{perfhue!55}
\colorlet{perf60}{perfhue!66}
\colorlet{perf70}{perfhue!77}
\colorlet{perf80}{perfhue!88}

\newcommand{\RN}[1]{%
	\textup{\lowercase\expandafter{\it \romannumeral#1}}%
}

\SetKwInput{KwInput}{Input}
\SetKwInput{KwOutput}{Output}
\DontPrintSemicolon

\SetKwComment{Comment}{\color{ggreen}\# }{}

\SetKwProg{Function}{def}{:}{}

\SetKwProg{For}{for}{:}{}
\SetKwProg{If}{if}{:}{}

\usepackage{needspace}
\usepackage{float}
\usepackage{placeins}
\usepackage{url}

\definecolor{easyppoaccent}{HTML}{B53146}
\hypersetup{linkcolor=easyppoaccent,citecolor=easyppoaccent,urlcolor=easyppoaccent}
\title{\texorpdfstring{Easy\textcolor{easyppoaccent}{PPO}}{EasyPPO}: Stabilizing the Critic Is Key}

\author[1,$*$,$\dagger$]{Xuanyi Zhou}
\author[1,$*$]{Qiuyang Mang}
\author[1,$*$]{Huanzhi Mao}
\author[1]{Dacheng Li}
\author[2]{Wenhao Chai}
\author[1]{Mayank~Mishra}
\author[1]{Yichuan~Wang}
\author[2]{Karthik Narasimhan}
\author[1]{Alvin Cheung}
\author[1]{Joseph~E.~Gonzalez}
\affil[1]{University of California, Berkeley}
\affil[2]{Princeton University}
\renewcommand{\institutionlogos}{%
  \includegraphics[height=30pt]{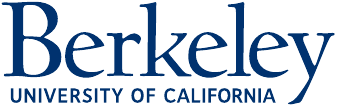}\hspace{18pt}%
  \includegraphics[height=28pt]{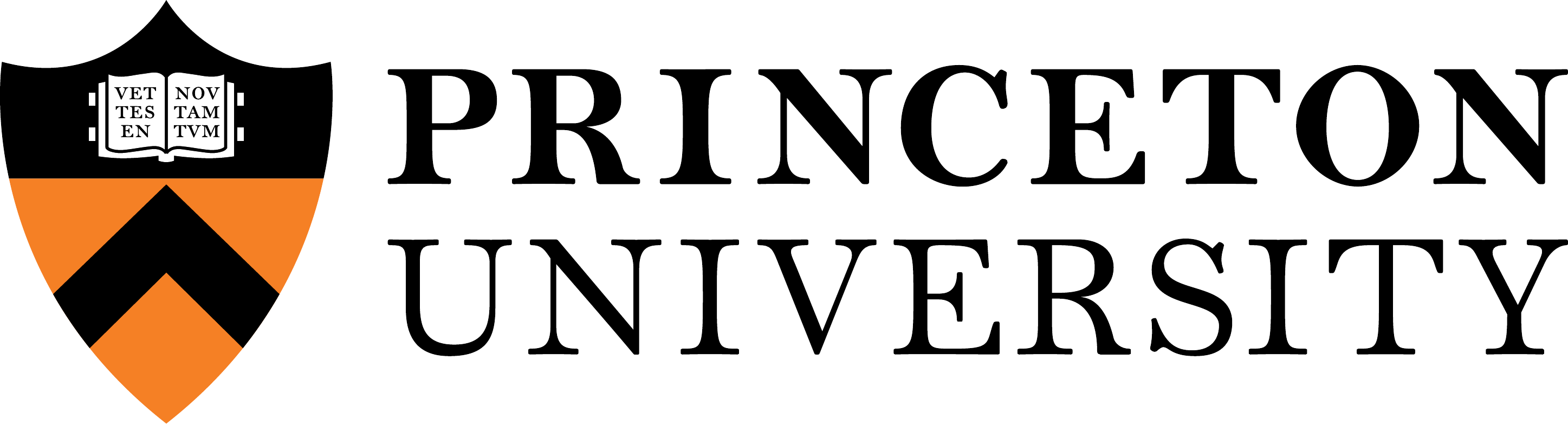}}
\authornote{$*$\,Equal contribution.\quad $\dagger$\,Project leader.\quad Correspondence to \texttt{qmang@berkeley.edu}.\\[5pt]
Website: \href{https://easyppo.github.io/}{\texttt{https://easyppo.github.io/}}\quad
Code: \href{https://github.com/EasyPPO/EasyPPO}{\texttt{https://github.com/EasyPPO/EasyPPO}}}

\begin{document}

\maketitle

\thispagestyle{firstpagestyle}

\begin{abstract}
{\rmfamily\bfseries\normalsize Abstract}\par\vspace{0.15\baselineskip}
\small
A key strength of Proximal Policy Optimization (PPO) is its learned critic, which uses historical trajectories collected during reinforcement learning to estimate expected returns and reduce policy-gradient variance.
However, we find that the critic is also a major source of instability in reinforcement learning for large language models (LLMs).
We identify two critic failure modes that destabilize PPO.
First, filtering truncated rollouts from both actor and critic shifts the policy objective to reward conditioned on completion, allowing truncation to increase even as conditional reward improves.
Second, heterogeneous return noise can cause high-variance prompts to dominate critic updates in finite batches.
We introduce \textbf{EasyPPO} to address these failures.
Actor-only overlong filtering trains the critic on returns from both completed and truncated rollouts.
Noise-normalized critic regression weights each prompt's critic loss by the inverse standard deviation of its sampled returns, balancing noise contributions across prompts.
Moderately smaller critic mini-batches confine outlier influence to fewer rollouts during gradient clipping.
Across continuous-reward coding on FrontierCS, binary-reward mathematical reasoning on AIME24, and multi-turn search on Search-R1, EasyPPO remains stable throughout the full training horizon and consistently outperforms vanilla PPO, VAPO, and HL-Gauss PPO.
Its best validation scores show relative gains of $14.89\%$, $2.28\%$, and $9.47\%$ over PPO, respectively.
\end{abstract}
\vspace{0.75em}

\section{Introduction}
\label{sec:intro}

\suppressfloats[t]   
\begin{figure}[t]
\centering
\includegraphics[width=\linewidth]{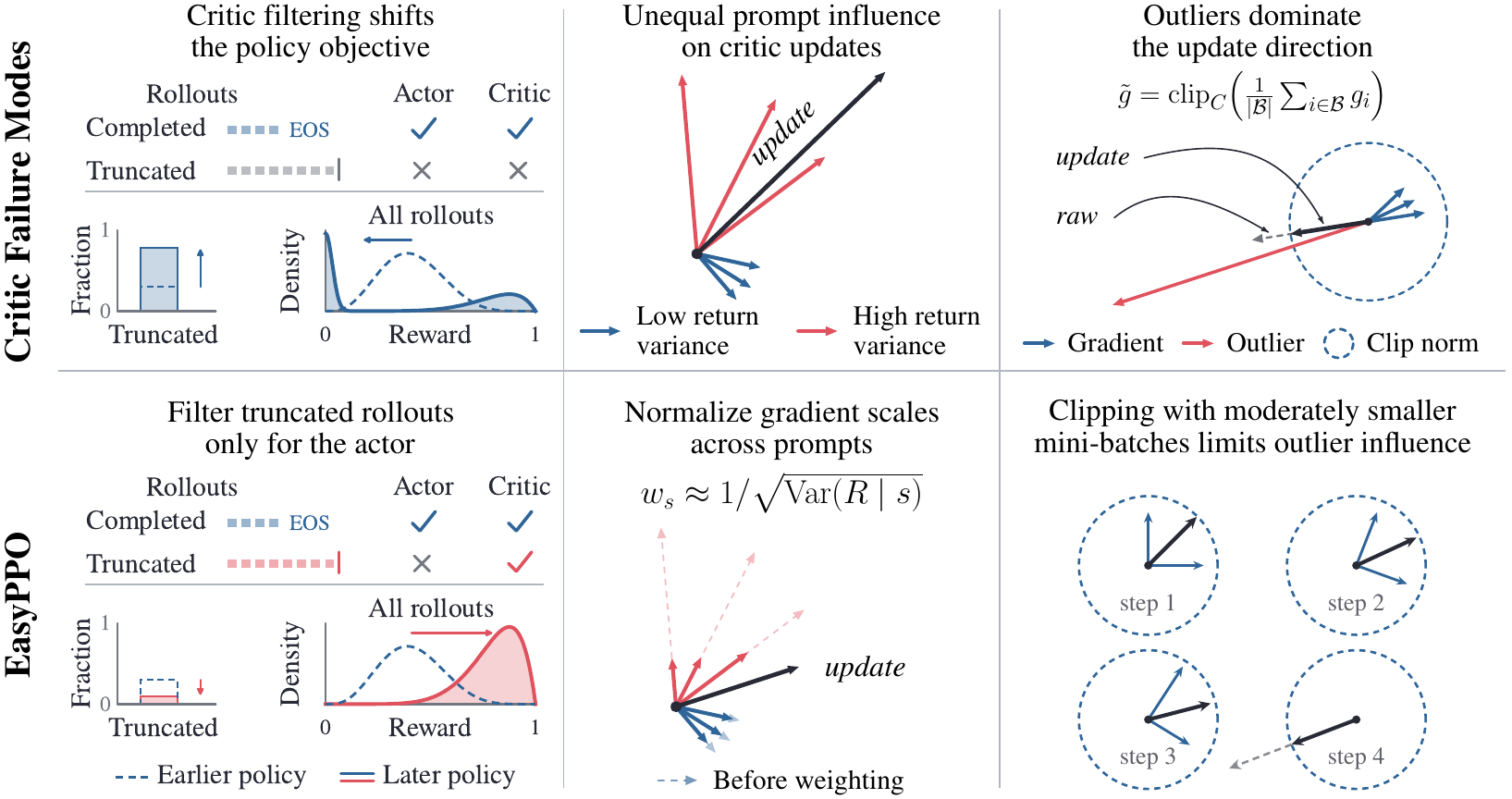}
\caption{\textbf{EasyPPO addresses two sources of critic instability with three simple modifications.}
\textbf{Left:} Filtering truncated rollouts from both actor and critic updates can increase truncation and lower overall reward; EasyPPO retains these rollouts for critic updates only.
\textbf{Middle:} Return variability differs across prompts, causing unequal critic gradient scales; EasyPPO weights prompts inversely by their standard deviation.
\textbf{Right:} Smaller critic mini-batches limit outlier influence through gradient clipping.}
\label{fig:teaser}
\end{figure}

Reinforcement learning with verifiable rewards (RLVR) has become a key post-training paradigm for improving reasoning in Large Language Models (LLMs)~\citep{openai2024o1,guo2025deepseekr1}.
Among RL algorithms, Proximal Policy Optimization (PPO) is particularly appealing for long-horizon reasoning because its actor--critic framework learns token-level return estimates from historical rollouts, enabling fine-grained credit assignment and lower-variance policy updates~\citep{schulman2017ppo}.
These benefits may become more important as post-training expands to tasks such as automated research~\citep{mang2025frontiercs,he2026frontiersmith,lyu2026mls,xu2026autolab,zhu2026edgebench}, where rollouts grow longer and require more computation on evaluation.

The critic can also become a source of instability in PPO.
Even with rollout batches of 512 responses in our experiments, we observe two failure modes in critic learning that heavily destabilize training.

\paragraph{Policy Objective Shift from Overlong Rollout Filtering.}
Truncated rollouts are a known source of reward noise in LLM reinforcement learning~\citep{yu2026dapo}.
Their rewards reflect incomplete responses, which can destabilize policy updates.
Overlong filtering mitigates this noise by excluding truncated rollouts from policy updates~\citep{yu2026dapo}.
For example, during the 24K stage of Math RL, Nemotron-Cascade~\citep{wang2025nemotroncascade} excludes overlong rollouts to avoid noisy penalties on unfinished reasoning.
In PPO, however, extending this filtering to critic updates can adversely change what the critic learns.
For a prompt $s$, the critic $V_\phi(s)$, parameterized by $\phi$, should predict the mean return $\mathbb{E}[R\mid s]$, where $R$ is the sampled rollout return.
Training it only on non-truncated rollouts instead makes it predict the mean return among completed responses.
We show that, in the accurate-critic limit, filtering both networks shifts the policy objective to reward conditioned on completion, $\mathbb{E}[R\mid s,\text{not truncated}]$.
This conditional reward can improve even as truncation becomes more frequent.
We observe a growing fraction of unfinished rollouts when filtering both actor and critic updates.
The critic should therefore retain these rollouts despite the noise in their returns.

\Needspace{3\baselineskip}
\paragraph{Heterogeneous Return Noise in Critic Updates.}
Retaining all rollouts for critic learning raises a second question: \emph{how should the critic handle return noise that varies across prompts within the same batch?}
These differences arise not only from truncation, but also because the actor policy produces consistent returns on some prompts and highly variable returns on others.
They are especially pronounced in continuous-reward tasks, where some prompts permit only small gains while others admit widely varying levels of improvement.
The critic aims to predict the mean return, minimizing $(V_\phi(s)-\mathbb{E}[R\mid s])^2$, but learns from sampled losses $(V_\phi(s)-R)^2$.
Our analysis shows that return noise leaves the optimal prediction unchanged, yet contributes to the expected squared gradient magnitude alongside prediction error.
As predictions improve, this noise can dominate, so prompts with more variable returns can disproportionately influence finite-batch updates.

Together, these two failures suggest that stable PPO should balance overlong filtering and critic learning while controlling how heterogeneous return noise shapes finite-batch updates.
We introduce \textbf{EasyPPO}, which addresses these failures through three simple yet effective modifications to PPO.
\Cref{fig:teaser} connects the two critic failure modes to the three changes in EasyPPO.
First, we apply \emph{actor-only overlong filtering}, retaining all rollouts for critic learning.
Second, \emph{noise-normalized critic regression} weights each prompt inversely by the empirical return standard deviation of its rollout group, to focus critic updates on prediction error normalized by return variability.
Third, although larger critic mini-batches better average out return noise, we found that smaller ones can be preferable under gradient clipping because they confine the remaining outliers' influence to fewer rollouts; we therefore choose a moderate size to balance these effects.

Our experiments across continuous-reward coding on FrontierCS~\citep{mang2025frontiercs,he2026frontiersmith}, binary-reward mathematical reasoning on AIME24, and multi-turn search on Search-R1~\citep{jin2025search} show that EasyPPO remains stable throughout 200--300 training updates, substantially improving training stability over vanilla PPO and two recent PPO variants, VAPO~\citep{yue2025vapo} and HL-Gauss PPO~\citep{zhou2026start}.
This stability translates into consistently stronger task performance, without the late-stage collapse frequently observed in the baselines.
EasyPPO's best validation scores show relative gains of $14.89\%$, $2.28\%$, and $9.47\%$ over PPO, respectively; it ranks first and is the only compared method stable across all three tasks.
Ablation studies further show that each component improves training stability and that combining all three yields the strongest gains.
A stable critic is therefore key to reliable PPO training for LLMs.

\section{Preliminaries}
We study PPO for reinforcement learning with verifiable rewards.
Given a prompt $x$, the policy $\pi_\theta$ generates a response $y_{1:T}$ and receives a scalar reward $R=R(x,y_{1:T})$ at the end of the rollout.
At token $t$, the state is the response prefix $s_t=(x,y_{<t})$ and the action is $a_t=y_t$.
PPO learns a critic $V_\phi(s_t)$ to estimate the expected return from each token state and uses these value estimates to construct token-level advantages with GAE~\citep{schulman2017ppo}.
In our terminal-reward setting, the return from any prefix is the final rollout reward $R$ when $\gamma=1$, so $V^{\pi_\theta}(s_t)=\mathbb{E}[R\mid s_t]$.

For simplicity, we treat a response $\tau$ as one action from prompt state $s$, with final reward $R$ and baseline $V_\phi(s)$.
We omit PPO clipping and KL regularization, yielding the actor objective
\begin{equation}
\mathcal{L}_{\mathrm{actor}}(\theta)
=
-\mathbb{E}_{\tau}\!\left[
\log\pi_\theta(\tau\mid s)
\operatorname{sg}\!\left(R-V_\phi(s)\right)
\right],
\label{eq:simplified-actor-loss}
\end{equation}
where $\mathbb{E}_\tau$ averages over sampled responses and $\operatorname{sg}(\cdot)$ stops gradients through the advantage.
The critic is trained against the final reward using the standard MSE objective
\begin{equation}
\mathcal{L}_{V}(\phi)
=
\frac{1}{2}\mathbb{E}_{\tau}\!\left[
\left(V_\phi(s)-R\right)^2
\right].
\label{eq:vanilla-critic-loss}
\end{equation}

Our implementation retains token-level GAE, the full clipped PPO objective, and KL regularization.
We consider a fully synchronized setting in which each rollout batch is sampled from the latest policy snapshot.

\section{Overlong Filtering in PPO}
\label{sec:actor-only-filter}

\begin{figure}[t!]
\centering
\includegraphics[width=\textwidth]{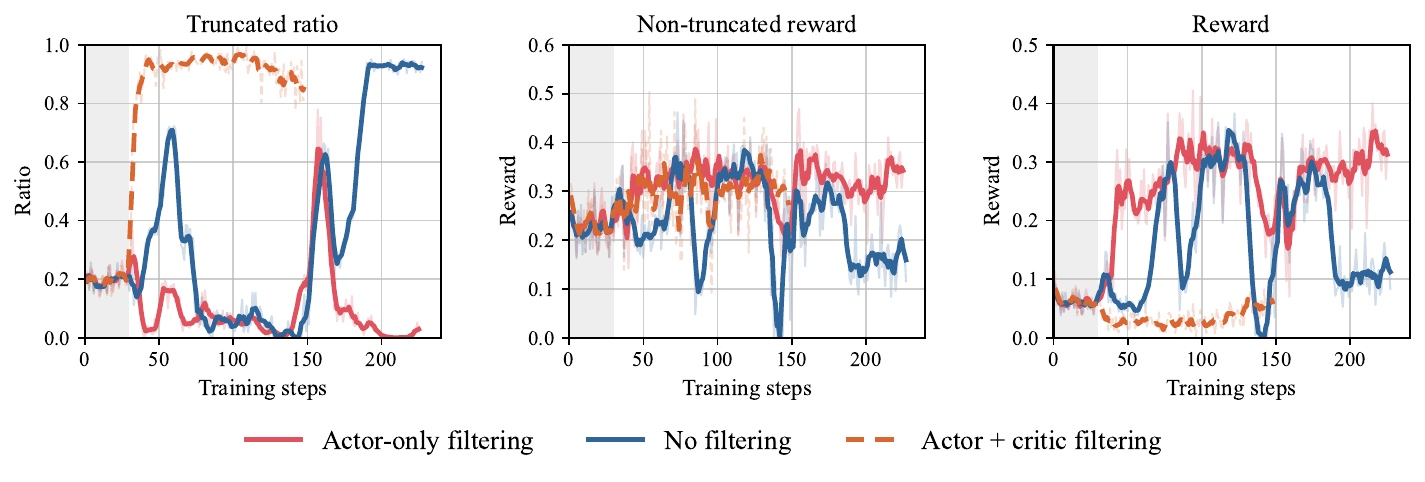}
\caption{
\textbf{Actor-only overlong filtering improves stability but does not eliminate instability.}
Rollout statistics during PPO training of Qwen3.5-9B in the FrontierCS setting, using $200$ problems generated by FrontierSmith, rollout batches of $512$, a $30$-step critic warm-up (gray), and a $32{,}768$-token response limit.
\textbf{Left:} fraction of truncated rollouts. \textbf{Middle:} mean reward among non-truncated rollouts. \textbf{Right:} mean reward over all rollouts.
\emph{Actor + critic filtering} improves reward among completed rollouts while the truncation ratio approaches $1$, leaving overall reward low.
\emph{No filtering} exhibits repeated reward collapses.
\emph{Actor-only filtering} achieves higher reward and lower truncation overall, but the spike and reward drop near step $160$ reveal residual instability.
}
\label{fig:frontiercs-overlong-filtering}
\end{figure}

Overlong filtering offers a straightforward way to reduce reward noise in Group Relative Policy Optimization (GRPO)~\citep{shao2024deepseekmath} by excluding truncated rollouts from policy updates~\citep{yu2026dapo,wang2025nemotroncascade}.
In PPO, however, filtering critic updates changes the learned value baseline and hence the advantages supplied to the actor.
\Cref{fig:frontiercs-overlong-filtering} compares three filtering strategies on FrontierCS~\citep{mang2025frontiercs}, using Qwen3.5-9B~\citep{qwen3.5} trained on $200$ problems generated by FrontierSmith~\citep{he2026frontiersmith}.
All runs use a supervised fine-tuning (SFT) checkpoint, a $30$-step critic warm-up, batches of $512$ rollouts, and a $32{,}768$-token response limit.

Without filtering, PPO exhibits repeated reward collapses.
Filtering both actor and critic updates improves reward among non-truncated rollouts, but the truncation ratio approaches $1$ and overall reward remains low.
Among the three strategies, actor-only filtering is the most stable and achieves the highest overall score.
Joint filtering also exhibits rising truncation and deteriorating overall reward in the AIME setting (\Cref{app:additional-filtering}).

\paragraph{Why Joint Filtering Shifts the Policy Objective.}
For a fixed prompt $s$, let $C$ denote the event that the rollout is not truncated.
Excluding truncated rollouts restricts the critic loss in \Cref{eq:vanilla-critic-loss} to completed responses, changing its pointwise population target from $\mathbb{E}[R\mid s]$ to $\mathbb{E}[R\mid s,C]$.

We analyze the idealized limit $V_\phi(s)=\mathbb{E}[R\mid s,C]$.
Let \mbox{$P_\theta(C\mid s)=\sum_{\tau\in C}\pi_\theta(\tau\mid s)>0$} be the completion probability.
Differentiating the conditional expected reward via the quotient rule gives
\begin{equation}
\begin{aligned}
\nabla_\theta\mathbb{E}[R\mid s,C]
&=\nabla_\theta
\frac{\sum_{\tau\in C}\pi_\theta(\tau\mid s)R}
{P_\theta(C\mid s)}\\[3pt]
&=\frac{\sum_{\tau\in C}\nabla_\theta\pi_\theta(\tau\mid s)R}
{P_\theta(C\mid s)}
-\underbrace{
\frac{\sum_{\tau\in C}\pi_\theta(\tau\mid s)R}{P_\theta(C\mid s)}
}_{V_\phi(s)\ =\  \mathbb{E}[R\mid s,C]}
\frac{\nabla_\theta P_\theta(C\mid s)}{P_\theta(C\mid s)}.
\end{aligned}
\label{eq:filtered-quotient-rule}
\end{equation}

\noindent Substituting \mbox{$\nabla_\theta P_\theta(C\mid s)=\sum_{\tau\in C}\nabla_\theta\pi_\theta(\tau\mid s)$} and applying the log-derivative identity gives
\begin{equation}
\begin{gathered}
\nabla_\theta\mathbb{E}[R\mid s,C]
=\frac{\sum_{\tau\in C}\nabla_\theta\pi_\theta(\tau\mid s)
\bigl(R-V_\phi(s)\bigr)}
{P_\theta(C\mid s)}\\[3pt]
=\sum_{\tau\in C}\frac{\pi_\theta(\tau\mid s)}{P_\theta(C\mid s)}
\bigl(R-V_\phi(s)\bigr)\nabla_\theta\log\pi_\theta(\tau\mid s)
=\underbrace{\mathbb{E}\!\left[
(R-V_\phi(s))\nabla_\theta\log\pi_\theta(\tau\mid s)
\,\middle|\,s,C
\right]}_{\textbf{Expected policy gradient of actor--critic filtering}}.
\end{gathered}
\label{eq:double-filter-objective}
\end{equation}
\Cref{eq:double-filter-objective} shows that joint filtering optimizes reward conditioned on completion, without directly encouraging the policy to avoid truncation.
This matches \Cref{fig:frontiercs-overlong-filtering}: reward among completed rollouts rises while truncation becomes more frequent and overall reward remains low.
At the token level, filtering similarly changes each critic target from $\mathbb{E}[R\mid s_t]$ to $\mathbb{E}[R\mid s_t,C]$, but the response-level policy-gradient identity need not hold (\Cref{app:overlong-filter-analysis}).

\paragraph{Actor-Only Overlong Filtering.}
We therefore apply overlong filtering only to actor updates and retain all rollouts for critic training.
As in GRPO, excluding truncated rollouts still biases the actor update.
Retaining all rollouts preserves the critic target $\mathbb{E}[R\mid s]$, restoring a completion-probability term in the idealized filtered policy gradient. When completed rollouts have higher expected returns than truncated ones, this term encourages completion.
However, the remaining truncation spike and reward drop in \Cref{fig:frontiercs-overlong-filtering} show that retaining all rollouts alone is not sufficient for stability. We next examine how heterogeneous return noise affects critic updates.
We defer a full analysis of actor-only filtering to \Cref{app:overlong-filter-analysis}.

\section{Stabilizing the Critic}
\label{sec:prompt-weighted-critic}

Following \Cref{sec:actor-only-filter}, we retain all rollouts for critic training.
However, the actor's sampled returns can vary for the same prompt, and retaining truncated rollouts can increase this \emph{return noise}.
The noise level varies across prompts in the same batch, reflecting differences in truncation rates and in return variability among completed rollouts.
This heterogeneity can be pronounced in continuous-score tasks with different reward scales.
For example, in GPU kernel optimization, GEMM solutions may achieve only around $1\times$ speedup over an optimized baseline~\citep{xing2026flashinfer}, whereas specialized operators can exceed $100\times$ over their respective library baselines~\citep{yang2026flashlib}.

\paragraph{Noise-Normalized Critic Regression.}

At a prefix $s$, consider a linear value head $V_\phi(s)=W^\top h_\phi(s)$ with critic features $h_\phi(s)$.
For a sampled return $R$, the MSE loss $\mathcal{L}_{V}(\phi)=\frac{1}{2}(V_\phi(s)-R)^2$ has gradient $\nabla_W\mathcal{L}_{V}(\phi)=(V_\phi(s)-R)h_\phi(s)$.
Since $R-\mathbb{E}[R\mid s]$ has zero conditional mean, the gradient second moment decomposes as
\begin{equation}
\label{eq:prompt-bias}
\mathbb{E}\!\left[
\left\|\nabla_W \mathcal{L}_{V}(\phi)\right\|_2^2
\,\middle|\, s
\right]
=
\left\|h_\phi(s)\right\|_2^2
\left[
\underbrace{
\bigl(V_\phi(s)-\mathbb{E}[R\mid s]\bigr)^2
}_{\textbf{Prediction error}}
+
\underbrace{
\operatorname{Var}(R\mid s)
}_{\textbf{Return noise}}
\right].
\end{equation}
As prediction error decreases, return noise can dominate this second moment, giving prompts with greater return variability disproportionate influence in finite batches.

This decomposition suggests dividing the critic loss at each prefix $s$ by its own conditional return standard deviation.
With $w(s)=1/\sqrt{\operatorname{Var}(R\mid s)}$ for positive conditional variance, \Cref{eq:prompt-bias} gives
\begin{equation}
\mathbb{E}\!\left[
\left\|w(s)\nabla_W\mathcal{L}_V(\phi)\right\|_2^2
\,\middle|\,s
\right]
=
\left\|h_\phi(s)\right\|_2^2
\left[
\underbrace{
\Biggl(\frac{V_\phi(s)-\mathbb{E}[R\mid s]}{\sqrt{\operatorname{Var}(R\mid s)}}\Biggr)^2
}_{\textbf{Normalized prediction error}}
+
\underbrace{
\vphantom{\bigl(\frac{V_\phi(s)-\mathbb{E}[R\mid s]}{\operatorname{Var}(R\mid s)}\bigr)^2}\frac{\operatorname{Var}(R\mid s)}{\operatorname{Var}(R\mid s)}
}_{\textbf{Constant noise term}}
\right].
\label{eq:weighted-gradient-decomposition}
\end{equation}
Here, positive state weighting preserves the pointwise population optimum $\mathbb{E}[R\mid s]$.
Apart from the feature norm, the gradient second moment depends only on normalized prediction error.
Fixing the noise term at one removes differences in return-noise scale as a source of imbalance across prompts.

In practice, estimating return variance separately at each prefix is costly, so we use the prompt's return standard deviation across its token states.
With bounded critic-feature norms, this yields a common upper bound on the noise contribution to the gradient second moment, averaged over prefixes.
We estimate $\hat{\sigma}(s)$ from rollout groups including truncated responses, and normalize weights to mean one over the prompt batch $\mathcal{B}$:
\begin{equation}
\begin{aligned}
w(s)&=
\frac{|\mathcal{B}|/\max\{\hat{\sigma}(s),\varepsilon\}}
{\sum_{s'\in\mathcal{B}}1/\max\{\hat{\sigma}(s'),\varepsilon\}},
\qquad \varepsilon>0,\\[4pt]
\mathcal{L}_V^{\mathrm{weighted}}(\phi)
&=\frac{1}{2\sum_i T_i}\sum\nolimits_i w(s_i)\sum\nolimits_{t=1}^{T_i}\bigl(V_\phi(s_{i,t})-R_i\bigr)^2.
\end{aligned}
\label{eq:prompt-weighted-loss}
\end{equation}
Here, response $i$ has prompt $s_i$, token states $s_{i,t}$, and $T_i$ valid tokens; the loss follows widely used token-level averaging~\citep{sheng2025hybridflow,yue2025vapo}.
For discrete rewards with range $\Delta$ and group size $n$, we propose the floor $\varepsilon=\Delta/(2\sqrt{n})$, where groups with identical returns receive about twice the weight of groups containing one maximum reward and $n-1$ minimum rewards.

For simplicity, we use the same rollout group to estimate weights and train the critic, which can introduce bias but works well in practice (\Cref{sec:experiments}).
When fewer rollouts per prompt are desired, alternatives such as offline profiling with online updates could provide variance estimates; we leave these alternatives to future work.
\Cref{app:prompt-weighting} details these approximations and the variance floor, including the continuous-reward setting.

\begin{figure}[t]
\centering
\includegraphics[width=\textwidth]{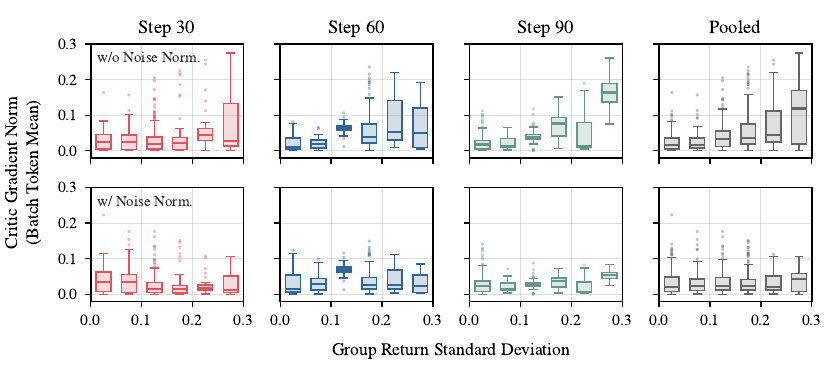}
\captionsetup{width=\textwidth}
\caption{\textbf{Noise normalization balances critic gradient contributions across prompts.}
Offline Qwen3.5-9B PPO diagnostics in the FrontierCS setting, using the data and sampling settings of \Cref{fig:frontiercs-overlong-filtering}.
\textbf{Columns:} steps $30$, $60$, $90$, and their pooled responses.
\textbf{Top:} without noise normalization. \textbf{Bottom:} with noise normalization.
Boxes summarize per-response gradient norms before clipping within each return-standard-deviation interval.
Consistent with \Cref{eq:prompt-bias}, prompts with more variable returns contribute larger gradients; noise normalization markedly reduces this dependence.}
\label{fig:critic-gradient-scatter}
\end{figure}

To test whether prompts with noisier returns contribute larger critic gradients, we study Qwen3.5-9B trained on FrontierSmith with actor-only filtering and noise-normalized critic regression.
We use the initialization and rollout settings of \Cref{fig:frontiercs-overlong-filtering}.
At training steps $30$, $60$, and $90$, we analyze $512$ responses per checkpoint ($16$ prompts, $32$ responses each).
Offline, we compute critic loss gradients for predicting each response's sampled return, sum them over its tokens, and divide by the batch's total token count.
\Cref{fig:critic-gradient-scatter} compares these gradient norms before clipping, with and without noise normalization, grouped by the empirical prompt return standard deviation.

Without normalization, prompts with more variable returns tend to contribute larger gradients, and this pattern is stronger at later checkpoints.
This is consistent with \Cref{eq:prompt-bias}: as the critic learns to predict the mean return, prediction error decreases, and return noise plays a larger role in the gradient second moment.
Noise normalization largely removes this dependence on return variability, making gradient contributions more balanced across prompts.
We also observe a similar overall trend in the AIME setting.
Note that some gradient outliers still remain, so we next adjust the critic mini-batch size to limit their impact through clipping.
\Cref{app:offline-gradient-diagnostics} provides the AIME results, scatter plots for both tasks at each checkpoint, and details of the gradient calculation.

\paragraph{Critic Mini-Batch Updates.}
\label{sec:minibatch-grad-clip}

The group estimates above can understate a prompt's return variability and give it excessive weight, leaving gradient outliers.
Even with exact weights, stochastic return noise remains.
Let $B$ and $m$ denote the numbers of rollouts in a critic batch and each mini-batch, respectively.
For each of the $\frac{B}{m}$ mini-batches, we compute the gradient $g$ of the mini-batch estimate of \Cref{eq:prompt-weighted-loss} and clip its norm to at most $c>0$ via $g\leftarrow g\cdot\min\{1,c/\|g\|_2\}$.
We take one optimizer step with this clipped gradient, then recompute the next mini-batch gradient at the updated critic parameters.

Smaller mini-batches confine the joint rescaling caused by outliers to fewer rollouts, but average out less return noise.
At fixed $B$, an idealized analysis gives an $O(m)$ bound on a single outlier's influence on the average clipped gradient, versus $O(1/m)$ gradient variance per mini-batch.
We defer the assumptions and proof to \Cref{app:minibatch-clipping}.
In practice, we use $\frac{B}{m} = 4$ critic mini-batches per rollout batch.
When changing the mini-batch size, the critic learning rate should be adjusted accordingly~\citep{li2026largerbatches}.

\begin{figure}[t]
\centering
\includegraphics[width=\linewidth]{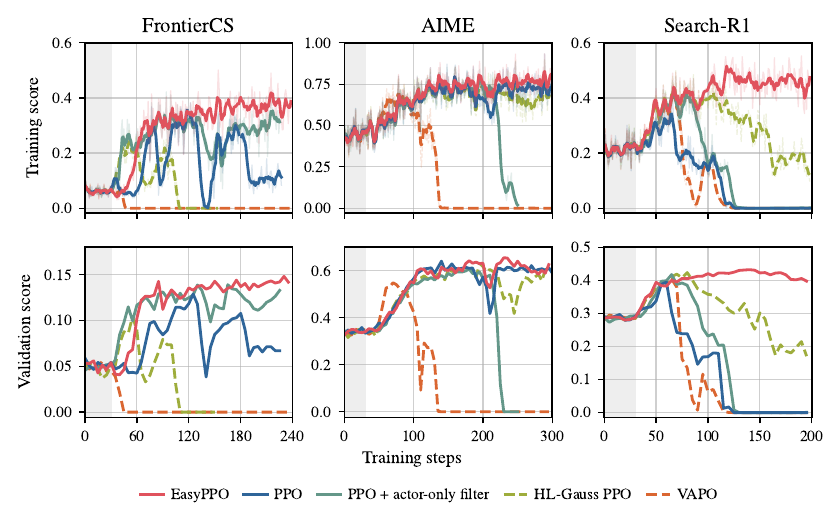}
\caption{\textbf{EasyPPO sustains learning across tasks.}
\textbf{Left:} FrontierCS. \textbf{Middle:} AIME. \textbf{Right:} Search-R1.
\textbf{Top:} training scores. \textbf{Bottom:} validation scores. Gray marks critic warmup.
}
\label{fig:main-results}
\end{figure}

\section{Experiments}
\label{sec:experiments}

\paragraph{Experimental Setup.}
For continuous-score coding, we train on a separate set of $200$ problems generated by FrontierSmith~\citep{he2026frontiersmith} and validate on the algorithmic track of FrontierCS~\citep{mang2025frontiercs}.
Their diversity lets us study cross-prompt return heterogeneity, complementing TailRL's focus on high-reward exploration~\citep{ramasubramanian2026taillikelihoodreinforcementlearning}.
To test stability on single-turn and multi-turn binary-reward tasks, we also train on DAPO-Math-17K~\citep{yu2026dapo} with AIME24 validation, and on the Search-R1 mixture with its seven validation datasets~\citep{jin2025search}.

We implement all methods in verl~\citep{sheng2025hybridflow}, using its vanilla PPO~\citep{schulman2017ppo} as the baseline.
We also compare with PPO + actor-only filtering, HL-Gauss PPO~\citep{zhou2026start} for its alternative critic objective, and VAPO~\citep{yue2025vapo} for its value-learning and advantage-estimation improvements.
We omit VAPO's auxiliary positive-example language-modeling loss to focus on PPO updates.
EasyPPO uses actor-only overlong filtering; PPO, HL-Gauss PPO, and VAPO use no filtering, following their original recipes.

We use Qwen3.5-9B-Base for AIME and Search-R1, and Qwen3.5-9B for FrontierCS~\citep{qwen3.5}.
For FrontierCS, we initialize from SFT on $347$ nonzero-score trajectories generated by DeepSeek-V3.1 on the same $200$ training problems.
All methods share a $30$-step critic warmup, rollout batches of $512$ responses on FrontierCS and AIME and $1024$ on Search-R1, and group sizes of $32$ on FrontierCS and $16$ on AIME and Search-R1.
Training is strictly on-policy, with actor updates starting after critic warmup.

\begin{figure}[t]
\centering
\includegraphics[width=0.8\linewidth]{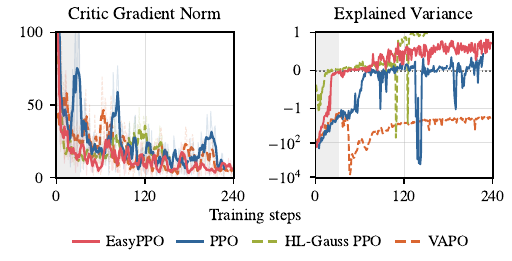}
\caption{\textbf{EasyPPO stabilizes critic learning.}
Critic diagnostics on FrontierCS.
\textbf{Left:} critic gradient norm before clipping.
\textbf{Right:} explained variance of critic value predictions.
Gray marks critic warmup.
}
\label{fig:fcs-critic-diagnostics}
\end{figure}

Training metrics are task score on FrontierSmith, reward on DAPO-Math-17K, and accuracy on Search-R1.
Validation averages over five responses per FrontierCS problem and $32$ per AIME24 problem; Search-R1 averages greedy accuracies equally across seven datasets and their own metrics.
Training retains the original DAPO-Math-17K rewards in $\{-1,+1\}$. For plotting only, we map these rewards to $(r+1)/2$ and divide FrontierCS scores by $100$.
\Cref{fig:main-results} shows one run per method and task.
Training curves use a five-point centered moving average with faint raw values; validation is unsmoothed.
Full configurations are provided in \Cref{app:hyperparameters}.

\paragraph{EasyPPO Stabilizes Critic Learning and Policy Training.}
Across the evaluated runs and training horizons in \Cref{fig:main-results}, EasyPPO remains stable and achieves the best validation performance among the compared methods on all three tasks.
Comparing each method's best validation checkpoint on a $0$ -- $100$ scale, EasyPPO improves over PPO by $1.92$, $1.46$, and $3.74$ points on FrontierCS, AIME24, and Search-R1, respectively; gains over the second-best method on each task are $0.91$, $1.46$, and $0.89$ points (\Cref{tab:best-validation-scores}).
The stability benefit is particularly pronounced on continuous-score coding, consistent with our analysis of critic instability under heterogeneous returns.
Every baseline experiences performance collapse in at least one setting, despite achieving substantial scores earlier in training.
With the EasyPPO recipe, we observe no training collapse in any of the evaluated settings.

PPO with actor-only filtering achieves higher FrontierCS scores than vanilla PPO, but its scores still fluctuate substantially. It also collapses on AIME. EasyPPO remains stable with the same filtering rule, showing the benefit of our critic modifications.

To examine the critic behavior behind these gains, \Cref{fig:fcs-critic-diagnostics} shows critic gradient norms before clipping and explained variance on FrontierCS.
After warmup, EasyPPO's critic gradient norm declines and its explained variance follows a steady upward trend with relatively small fluctuations, whereas the baselines exhibit sharp swings in explained variance.
HL-Gauss approaches an explained variance of $1$ only after its actor collapses.
In contrast, EasyPPO's improving critic predictions accompany sustained reward gains, supporting our motivation for stabilizing critic learning to improve PPO training.
We defer AIME and Search-R1 critic diagnostics to \Cref{app:additional-critic-diagnostics}.

\begingroup
\setlength{\intextsep}{4pt}
\setlength{\columnsep}{14pt}
\begin{wrapfigure}{r}{0.45\linewidth}
\vspace{-6pt}
\centering
\captionsetup{width=\linewidth,justification=centering,skip=4pt}
\includegraphics[width=\linewidth]{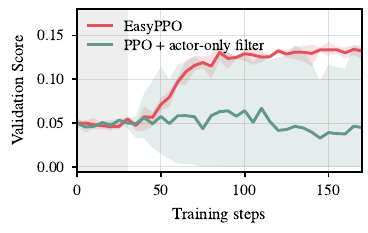}\par
\caption{\textbf{Stability across critic \mbox{initializations}.}}
\label{fig:fcs-seed-comparison}
\end{wrapfigure}

\begin{figure}[t]
\centering
\includegraphics[width=0.8\linewidth]{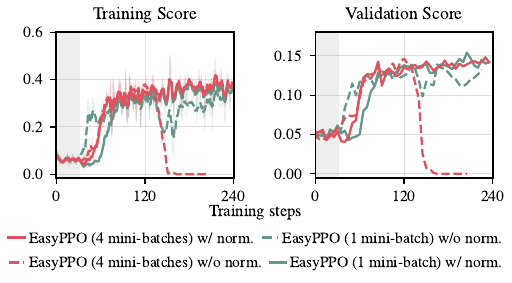}
\caption{\textbf{Noise normalization stabilizes training with various critic mini-batch configurations.}
FrontierCS ablation.
\textbf{Left:} training score. \textbf{Right:} validation score.
}
\label{fig:fcs-normalization-ablation}
\end{figure}

\paragraph{Sensitivity to Random Seeds.}
To test seed sensitivity, we compare three runs each of EasyPPO and PPO + actor-only filtering, the strongest baseline on FrontierCS.
Due to training cost, we restrict this study to these two methods on FrontierCS.
None of the three EasyPPO runs collapses, whereas two of the three actor-only-filtering runs do.
EasyPPO also sustains its validation gains with a narrow min--max band across runs (\Cref{fig:fcs-seed-comparison}).
Further details are provided in \Cref{app:fcs-repeated-runs}.
\par
\ifnum\value{WF@wrappedlines}>2
  \vspace{\dimexpr\value{WF@wrappedlines}\baselineskip-2\baselineskip-12pt\relax}
\fi
\WFclear
\endgroup

\vspace{1ex}
\paragraph{Noise Normalization Improves Stability Across Critic Mini-Batch Sizes.}
We ablate noise-normalized critic regression on FrontierCS.
Within each mini-batch configuration, all other training settings remain unchanged.
We use a fixed critic learning rate of $2\times10^{-6}$ for this ablation.

\Cref{fig:fcs-normalization-ablation} shows that noise normalization stabilizes training with both mini-batch configurations.

Without normalization, the performance drop is recoverable with one mini-batch but becomes a sustained collapse when the same rollout batch is split into four smaller mini-batches.
With normalization, both configurations preserve their gains in training and validation scores.
This contrast is consistent with our analysis: smaller mini-batches average out less return noise.
These results suggest that noise normalization enables stable training with smaller critic mini-batches, allowing finer-grained clipping to limit outlier influence as analyzed in \Cref{sec:minibatch-grad-clip}.

\begin{figure}[t]
\centering
\includegraphics[width=0.8\linewidth]{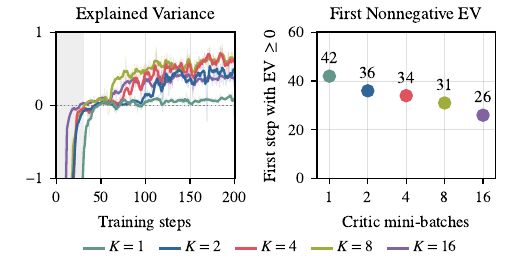}
\caption{\textbf{Explained variance across critic mini-batch sizes.}
FrontierCS. \textbf{Left:} smoothed explained variance; gray marks critic warmup.
\textbf{Right:} first training step with $\mathrm{EV}\geq0$ versus critic mini-batch count, measured from unsmoothed values.
}
\label{fig:fcs-minibatch-ablation}
\end{figure}

\paragraph{The Trade-off in Critic Mini-Batch Size.}

We next vary the number of critic mini-batches per rollout batch, $K=B/m$, on FrontierCS, fixing $B=512$ and retaining actor-only filtering and noise normalization.
We scale $\eta\propto1/\sqrt{K}$ from the default at $K=4$ to account for update frequency~\citep{li2026largerbatches}; clipping granularity and optimizer dynamics still vary together, but the observed trade-off is consistent with our fixed-parameter analysis.

{\Cref{fig:fcs-minibatch-ablation} shows five-point-smoothed EV (left) and the first step with raw $\mathrm{EV}\geq0$ (right), including warmup.
First-crossing steps decrease approximately linearly with $\log K$, whereas $K=4$ and $8$ achieve higher EV later in training.
This suggests finer-grained outlier control helps early, while noise averaging matters more as prediction error decreases (\Cref{eq:prompt-bias}).
Together, this analysis and the observed trade-off motivate our default choice of $K=4$.}

\section{Related Work}
\label{sec:related}

\paragraph{Policy Optimization for LLM Post-Training.}
Reinforcement learning from human feedback (RLHF) established PPO as a standard actor--critic method for LLM post-training~\citep{ouyang2022training,schulman2017ppo}, while RLVR replaces learned reward models with task-specific verifiers~\citep{shao2024deepseekmath,guo2025deepseekr1}.

Critic-free methods estimate advantages from groups of responses to the same prompt~\citep{shao2024deepseekmath,ahmadian2024back,hu2025reinforcepp,he2025justrl},
commonly filtering truncated rollouts from the update~\citep{yu2026dapo,wang2025nemotroncascade}.

Actor--critic methods keep a critic for token-level credit assignment, and recent work improves its training through value pretraining and decoupled GAE~\citep{yuan2025vcppo,yue2025vapo}, categorical value prediction~\citep{zhou2026start}, or hybrid designs~\citep{qi2026bpco,pan2026evpo}.
Concurrent work supervises the critic sparsely to counter value flattening inside a response~\citep{li2026valueflattening},
takes multiple critic updates per rollout batch~\citep{hu2025openreasonerzero},
excludes length-penalty rewards from critic training~\citep{justrl2},
or reweights actor baselines by length~\citep{cognition2026swe2}.
Each addresses one piece of critic training; EasyPPO targets the critic's exposure to prompt-level return noise, showing that filtering both actor and critic shifts the policy objective, normalizing critic-gradient noise across prompts, and bounding each mini-batch's influence.

\paragraph{Continuous Verifiable Rewards.}
While mathematical and coding RLVR often use binary correctness rewards, open-ended optimization admits continuously varying solution quality.
Recent benchmarks provide verifiable graded scores for algorithm design, ML research, and GPU kernel optimization~\citep{mang2025frontiercs,kong2026frontieror,he2026frontiersmith,zhu2026edgebench,lyu2026mls,xing2026flashinfer}.
These benchmarks also provide training data: Evolution Fine-Tuning uses evolutionary search trajectories from FrontierCS for supervised fine-tuning~\citep{lee2026evolution}.
For RL with continuous rewards, TailRL~\citep{ramasubramanian2026taillikelihoodreinforcementlearning} targets upper-tail outcomes through reward-threshold exceedance probabilities based on GRPO.
These settings make differences in return scale and variability across prompts particularly visible, motivating EasyPPO's focus on critic stability under heterogeneous returns.

\paragraph{Variance-Weighted Critic Regression.}
Variance-aware regression has been studied in both RL and supervised learning.
In RL, variance-weighted critic objectives improve statistical efficiency in linear MDPs~\citep{zhou2021nearly,kitamura2023regularization}, while IV-RL weights TD errors using estimated target variance~\citep{mai2022sample} and PopArt normalizes target scales across tasks~\citep{van2016learning,hessel2019multi}.
Related work on heteroscedastic regression similarly adjusts each example's contribution according to its target uncertainty~\citep{kendall2017uncertainties,seitzer2022pitfalls}.
Group sampling in LLM RL provides an empirical estimate of prompt-level return variance, which EasyPPO uses for noise-normalized critic regression without an auxiliary variance model.
Dr.~GRPO finds that standard-deviation normalization of actor advantages can over-weight low-variance prompts~\citep{liu2025understanding}; EasyPPO instead applies this weighting to the critic loss.

\section{Conclusion}
\label{sec:conclusion}
We identify overlong-rollout handling and heterogeneous return noise as two sources of critic instability in PPO.
Our analysis shows how filtering both actor and critic changes the policy objective, and how return noise can dominate critic-gradient second moments.
These findings motivate EasyPPO: actor-only filtering, noise-normalized critic regression, and moderately smaller critic mini-batches with gradient clipping, while retaining the standard PPO actor update.
Across continuous-score coding, mathematical reasoning, and multi-turn search, the resulting stability and performance gains show that stabilizing the critic is key to reliable PPO training for LLMs.

\section*{Acknowledgments}
We thank the Laude Institute and Modal, as well as Ziniu Li, Yiping Wang, Shuning Shang, Shuo Yang, Bo Peng, Haocheng Xi, Runyuan He, Kaiyuan Liu, Yi Pan, Shuo Yuan, and Peter Chen, for supporting us and discussing this paper.

\begingroup
\setlength{\bibsep}{1pt plus 0.5pt}
\bibliography{references}
\endgroup

\clearpage
\beginsupplement
\crefalias{section}{appendix}   
\raggedbottom
\setcounter{topnumber}{2}
\setcounter{totalnumber}{4}
\renewcommand{\topfraction}{0.78}
\renewcommand{\bottomfraction}{0.95}
\renewcommand{\textfraction}{0.20}
\makeatletter
\setlength{\@fptop}{0pt}
\setlength{\@fpsep}{16pt}
\setlength{\@fpbot}{0pt plus 1fil}
\makeatother

\section*{Appendix Contents}
\phantomsection
\label{sec:appendix}

\newcommand{\appendixentry}[2]{%
  \noindent\hspace*{#1}\hyperref[#2]{\ref*{#2}\quad\nameref*{#2}}%
  \nobreak\dotfill\hyperref[#2]{\pageref*{#2}}\par}
\begingroup
\small
\setlength{\parskip}{2pt}
\hypersetup{linkcolor=black}
{\bfseries\appendixentry{0pt}{app:hyperparameters}}
\appendixentry{1em}{app:distillation}
\appendixentry{1em}{app:training-config}
\appendixentry{1em}{app:critic-config}
\appendixentry{1em}{app:ablation-config}
\smallskip
{\bfseries\appendixentry{0pt}{app:theory}}
\appendixentry{1em}{app:return-variability}
\appendixentry{1em}{app:overlong-filter-analysis}
\appendixentry{1em}{app:prompt-weighting}
\appendixentry{1em}{app:minibatch-clipping}
\smallskip
{\bfseries\appendixentry{0pt}{app:additional-results}}
\appendixentry{1em}{app:best-validation-scores}
\appendixentry{1em}{app:additional-critic-diagnostics}
\appendixentry{1em}{app:fcs-repeated-runs}
\appendixentry{1em}{app:offline-gradient-diagnostics}
\appendixentry{1em}{app:additional-filtering}
\appendixentry{1em}{app:search-r1-validation-sets}
\endgroup

\section{Training Configurations and Implementation Details}
\label{app:hyperparameters}
This section provides the configurations for \Cref{sec:experiments} and the filtering study in \Cref{sec:actor-only-filter}.

\subsection{Data and FrontierCS Initialization}
\label{app:distillation}
For our FrontierCS experiments, we use a distilled Qwen3.5-9B~\citep{qwen3.5} model as the initial policy.
SFT and RL use a separate FrontierSmith training set; validation uses the algorithmic track of FrontierCS~\citep{mang2025frontiercs}.
We use DeepSeek-V3.1 to sample three responses for each of the $200$ problems generated by FrontierSmith~\citep{he2026frontiersmith}, yielding $600$ responses in total.
Each response is truncated to at most $32{,}768$ tokens, with length measured using the Qwen3.5-9B tokenizer.
After removing $253$ responses with a score of zero, we retain $347$ responses covering $179$ distinct problems.
We perform supervised fine-tuning (SFT) of Qwen3.5-9B on these retained responses for a single epoch to obtain the distilled model.

\subsection{Training and Evaluation Settings}
AIME and Search-R1 start from Qwen3.5-9B-Base; FrontierCS uses the SFT initialization above.
\label{app:training-config}
\Cref{tab:task-configuration} summarizes the task settings for \Cref{fig:main-results}.
Batch sizes count responses; the number of prompts per batch is listed separately.
\Cref{tab:ppo-configuration} lists the PPO and optimizer settings; method-specific critic settings follow in \Cref{app:critic-config}.
Learning rates are constant after any optimizer warmup.

\begin{table}[!htb]
\centering
\caption{Training and evaluation configurations for the main comparisons. Learning rates are the configured target values.}
\label{tab:task-configuration}
\setlength{\tabcolsep}{6pt}
\renewcommand{\arraystretch}{1.04}
\begin{tabularx}{0.92\linewidth}{@{}>{\raggedright\arraybackslash}X*{3}{>{\centering\arraybackslash}p{0.145\linewidth}}@{}}
\toprule
\rowcolor{gblue!6}
\textbf{Hyperparameter} & \textbf{FrontierCS} & \textbf{AIME} & \textbf{Search-R1} \\
\midrule
Prompts per rollout batch & 16 & 32 & 64 \\
Samples per prompt & 32 & 16 & 16 \\
Rollout batch size $B$ & 512 & 512 & 1024 \\
Maximum prompt length & 8192 & 2048 & 4096 \\
Maximum response length & 32768 & 8192 & 4096 \\
Rollout temperature & 1.0 & 1.0 & 1.0 \\
Rollout top-$p$ & 1.0 & 1.0 & 1.0 \\
\midrule
Actor learning rate & $10^{-6}$ & $10^{-6}$ & $10^{-6}$ \\
Default critic learning rate & $2\times10^{-6}$ & $2\times10^{-6}$ & $2\times10^{-6}$ \\
Actor updates per rollout batch & 1 & 1 & 1 \\
Critic epochs per rollout batch & 1 & 1 & 1 \\
Actor/critic LR warmup (optimizer steps) & 0 & 20 & 0 \\
Critic warmup steps & 30 & 30 & 30 \\
\midrule
Validation responses per problem & 5 & 32 & 1 \\
Validation decoding & Sampling & Sampling & Greedy \\
Validation temperature & 1.0 & 1.0 & 0.0 \\
Validation top-$p$ & 1.0 & 0.7 & 1.0 \\
\bottomrule
\end{tabularx}
\end{table}

Critic warmup is reported in rollout steps; the optimizer learning-rate warmup is listed separately.

\begin{table}[!htb]
\centering
\caption{PPO and optimizer settings. The Search-R1 upper clipping parameter is $0.28$ for VAPO and $0.2$ for the other methods.}
\label{tab:ppo-configuration}
\renewcommand{\arraystretch}{1.04}
\begin{tabularx}{0.92\linewidth}{@{}>{\raggedright\arraybackslash}X*{3}{>{\centering\arraybackslash}p{0.145\linewidth}}@{}}
\toprule
\rowcolor{gblue!6}
\textbf{Hyperparameter} & \textbf{FrontierCS} & \textbf{AIME} & \textbf{Search-R1} \\
\midrule
Discount factor $\gamma$ & 1 & 1 & 1 \\
GAE parameter $\lambda$ & 1 & 1 & 1 \\
PPO lower clipping parameter & 0.2 & 0.2 & 0.2 \\
PPO upper clipping parameter & 0.2 & 0.28 & 0.2 / 0.28 \\
Dual-clip parameter & 3 & 3 & 3 \\
MSE value-clipping parameter & 0.2 & 0.2 & 0.2 \\
Actor / critic gradient clipping & 1 / 1 & 1 / 1 & 1 / 1 \\
Actor / critic PPO epochs & 1 / 1 & 1 / 1 & 1 / 1 \\
Actor entropy coefficient & 0 & 0 & 0 \\
Actor KL-loss coefficient & $10^{-3}$ & $10^{-3}$ & $10^{-3}$ \\
KL penalty in reward & Off & Off & Off \\
Actor / critic loss aggregation & Token mean & Token mean & Token mean \\
\midrule
Optimizer & AdamW & AdamW & AdamW \\
AdamW $(\beta_1,\beta_2)$ & $(0.9,0.999)$ & $(0.9,0.999)$ & $(0.9,0.999)$ \\
Weight decay & 0.01 & 0.01 & 0.01 \\
AdamW numerical epsilon & 1e-8 & 1e-8 & 1e-8 \\
\bottomrule
\end{tabularx}
\end{table}
The actor KL loss uses the \texttt{low\_var\_kl} estimator.

\subsection{Critic and Baseline Configurations}
\label{app:critic-config}
\Cref{tab:method-configuration} summarizes the critic configurations in \Cref{fig:main-results}.
EasyPPO uses $K=4$ critic mini-batches per rollout batch, each with $m=B/K$ responses: $128$ on FrontierCS and AIME, and $256$ on Search-R1.
\Cref{tab:easyppo-configuration} lists the EasyPPO-specific settings.
All methods retain truncated responses for critic training.
VAPO retains its value-learning and advantage-estimation recipe, while its auxiliary positive-example language-modeling loss is disabled as described in \Cref{sec:experiments}.

\begin{table}[!htb]
\centering
\caption{Critic objectives, overlong filtering, and update counts in the main comparisons. $K$ is the number of critic mini-batches per rollout batch.}
\label{tab:method-configuration}
\setlength{\tabcolsep}{6pt}
\renewcommand{\arraystretch}{1.04}
\begin{tabularx}{0.92\linewidth}{@{}p{0.30\linewidth}X>{\centering\arraybackslash}p{0.19\linewidth}>{\centering\arraybackslash}p{0.04\linewidth}@{}}
\toprule
\rowcolor{gblue!6}
\textbf{Method} & \textbf{Critic objective} & \textbf{Overlong filtering} & \textbf{$K$} \\
\midrule
PPO & MSE & None & 1 \\
PPO + actor-only filtering & MSE & Actor only & 1 \\
HL-Gauss PPO & HL-Gauss & None & 1 \\
VAPO & MSE & None & 1 \\
EasyPPO & Noise-normalized MSE & Actor only & 4 \\
\bottomrule
\end{tabularx}
\end{table}

\begin{table}[!htb]
\centering
\caption{EasyPPO-specific configurations. Shared PPO settings appear in \Cref{tab:ppo-configuration}.}
\label{tab:easyppo-configuration}
\renewcommand{\arraystretch}{1.04}
\begin{tabularx}{0.92\linewidth}{@{}>{\raggedright\arraybackslash}X*{3}{>{\centering\arraybackslash}p{0.145\linewidth}}@{}}
\toprule
\rowcolor{gblue!6}
\textbf{Hyperparameter} & \textbf{FrontierCS} & \textbf{AIME} & \textbf{Search-R1} \\
\midrule
Critic mini-batches $K$ & 4 & 4 & 4 \\
Responses per critic mini-batch $m$ & 128 & 128 & 256 \\
Critic learning rate & $2\times10^{-6}$ & $2\times10^{-6}$ & $2\times10^{-6}$ \\
Overlong filtering & Actor only & Actor only & Actor only \\
\midrule
Return-STD floor $\varepsilon$ & 0.075 & 0.25 & 0.125 \\
\bottomrule
\end{tabularx}
\end{table}


\Cref{tab:hl-gauss-configuration} gives the task-specific categorical-critic settings used by our HL-Gauss PPO baseline~\citep{zhou2026start}.
\begin{table}[!htb]
\centering
\caption{HL-Gauss critic configurations used in our experiments.}
\label{tab:hl-gauss-configuration}
\renewcommand{\arraystretch}{1.04}
\begin{tabularx}{0.92\linewidth}{@{}>{\raggedright\arraybackslash}X*{3}{>{\centering\arraybackslash}p{0.145\linewidth}}@{}}
\toprule
\rowcolor{gblue!6}
\textbf{Hyperparameter} & \textbf{FrontierCS} & \textbf{AIME} & \textbf{Search-R1} \\
\midrule
Number of bins & 101 & 101 & 101 \\
Support range & $[-0.1,1.1]$ & $[-1.1,1.1]$ & $[-0.1,1.1]$ \\
Smoothing bandwidth & $0.009$ & $0.0165$ & $0.009$ \\
\bottomrule
\end{tabularx}
\end{table}

\subsection{Diagnostic and Ablation Configurations}
\label{app:ablation-config}
\paragraph{Overlong filtering.}
For \Cref{fig:frontiercs-overlong-filtering}, we use Qwen3.5-9B~\citep{qwen3.5} and train on $200$ problems generated by FrontierSmith~\citep{he2026frontiersmith}.
We use the distilled initialization described in \Cref{app:distillation}.
We warm up the critic for $30$ steps before updating the policy with PPO, using rollout batches of $512$.
The maximum response length is $32{,}768$ tokens.
The three runs compare no overlong filtering, filtering both actor and critic updates, and filtering actor updates only.
The figure reports training-rollout statistics; the algorithmic track of FrontierCS~\citep{mang2025frontiercs} is used for validation.

\paragraph{Return-variability diagnostics.}
\Cref{tab:diagnostic-configuration} summarizes the shared settings and differences between \Cref{fig:frontiercs-overlong-filtering,fig:critic-gradient-scatter}.
\Cref{fig:critic-gradient-scatter} uses checkpoints from an actor-only-filtered, noise-normalized training run, and compares the same offline response gradients before and after weighting.
Its diagnostic return-STD floor is $\varepsilon=0.075$; the corresponding AIME diagnostic uses $0.5$.
The diagnostic uses terminal-return MSE gradients before clipping, without taking an optimizer step; its weighting and token denominator are detailed in \Cref{app:offline-gradient-diagnostics}.

\begin{table}[!htb]
\centering
\caption{FrontierCS settings for the filtering study and offline gradient diagnostic. Both use the distilled initialization from \Cref{app:distillation}, $512$ responses per rollout batch, group size $32$, a $32{,}768$-token response limit, and $30$ critic warmup steps.}
\label{tab:diagnostic-configuration}
\renewcommand{\arraystretch}{1.04}
\begin{tabularx}{0.92\linewidth}{@{}>{\raggedright\arraybackslash}X>{\centering\arraybackslash}p{0.25\linewidth}>{\centering\arraybackslash}p{0.25\linewidth}@{}}
\toprule
\rowcolor{gblue!6}
\textbf{Setting} & \textbf{\Cref{fig:frontiercs-overlong-filtering}} & \textbf{\Cref{fig:critic-gradient-scatter}} \\
\midrule
Actor learning rate & $10^{-6}$ & $10^{-6}$ \\
Critic learning rate & $2\times10^{-6}$ & $2\times10^{-6}$ \\
Critic mini-batches per rollout batch & 1 & 4 \\
Noise normalization during training & Off & On \\
Overlong filtering during training & None / both / actor only & Actor only \\
Diagnostic checkpoints & --- & 30, 60, 90 \\
Responses analyzed per checkpoint & --- & 512 \\
Diagnostic return-STD floor & --- & 0.075 \\
\bottomrule
\end{tabularx}
\end{table}

\paragraph{Noise normalization and critic mini-batches.}
\Cref{tab:ablation-configuration} lists the configurations plotted in \Cref{fig:fcs-normalization-ablation,fig:fcs-minibatch-ablation}.
All use FrontierCS with $B=512$, actor-only overlong filtering, and the initialization in \Cref{app:distillation}.
\Cref{fig:fcs-normalization-ablation} holds the critic learning rate fixed to isolate normalization within each mini-batch configuration.
\Cref{fig:fcs-minibatch-ablation} instead scales it as $\eta\propto1/\sqrt{K}$ from the default $K=4$ setting.

\begin{table}[!htb]
\centering
\caption{Configurations used in the FrontierCS ablation figures. $m=512/K$ counts responses per critic mini-batch. Only plotted configurations are listed.}
\label{tab:ablation-configuration}
\renewcommand{\arraystretch}{1.04}
\begin{tabularx}{0.92\linewidth}{@{}>{\raggedright\arraybackslash}X>{\centering\arraybackslash}p{0.05\linewidth}>{\centering\arraybackslash}p{0.08\linewidth}>{\centering\arraybackslash}p{0.25\linewidth}>{\centering\arraybackslash}p{0.25\linewidth}@{}}
\toprule
\rowcolor{gblue!6}
\textbf{Experiment} & \textbf{$K$} & \textbf{$m$} & \textbf{Noise normalization} & \textbf{Critic learning rate} \\
\midrule
\Cref{fig:fcs-normalization-ablation} & 1 & 512 & On / off & $2\times10^{-6}$ \\
\Cref{fig:fcs-normalization-ablation} & 4 & 128 & On / off & $2\times10^{-6}$ \\
\midrule
\Cref{fig:fcs-minibatch-ablation} & 1 & 512 & On & $4\times10^{-6}$ \\
\Cref{fig:fcs-minibatch-ablation} & 2 & 256 & On & $2\sqrt{2}\times10^{-6}$ \\
\Cref{fig:fcs-minibatch-ablation} & 4 & 128 & On & $2\times10^{-6}$ \\
\Cref{fig:fcs-minibatch-ablation} & 8 & 64 & On & $\sqrt{2}\times10^{-6}$ \\
\Cref{fig:fcs-minibatch-ablation} & 16 & 32 & On & $1\times10^{-6}$ \\
\bottomrule
\end{tabularx}
\end{table}

\FloatBarrier
\section{Theoretical Analyses}
\label{app:theory}

\subsection{Effect of Return Heterogeneity on Critic Updates}
\label{app:return-variability}
This derivation supports the gradient second-moment decomposition in \Cref{eq:prompt-bias}.

We provide a simple qualitative analysis to illustrate how
heterogeneous return noise affects critic optimization.

Let the critic be parameterized as
\begin{equation}
    V_{\phi}(s)
    =
    W^\top h_\phi(s),
\end{equation}
where \(h_\phi(s)\in\mathbb{R}^d\) denotes the hidden representation
produced by the critic backbone and \(W\in\mathbb{R}^d\) is the linear
value head.
Given a sampled rollout return \(R\), the critic minimizes the squared
value loss
\begin{equation}
    \ell
    =
    \frac{1}{2}
    \left(
        V_{\phi}(s)-R
    \right)^2.
    \label{eq:app-value-loss}
\end{equation}

The gradient with respect to the value head is
\begin{align}
    \nabla_W \ell
    &=
    \left(
        V_{\phi}(s)-R
    \right)
    \nabla_W V_{\phi}(s)
    \nonumber\\
    &=
    \left(
        V_{\phi}(s)-R
    \right)
    h_\phi(s).
    \label{eq:app-value-head-grad}
\end{align}

For a fixed prompt \(s\), define the conditional mean and variance of
the rollout return as
\begin{equation}
    \mu_s
    \triangleq
    \mathbb{E}[R\mid s],
    \qquad
    \sigma_s^2
    \triangleq
    \operatorname{Var}(R\mid s).
    \label{eq:app-return-stats}
\end{equation}

We first consider the expected value-head gradient.
Taking expectation over rollout returns conditioned on \(s\) gives
\begin{align}
    \mathbb{E}
    \left[
        \nabla_W \ell
        \mid s
    \right]
    &=
    \mathbb{E}
    \left[
        \left(
            V_{\phi}(s)-R
        \right)
        h_\phi(s)
        \mid s
    \right]
    \nonumber\\
    &=
    \left(
        V_{\phi}(s)-\mu_s
    \right)
    h_\phi(s).
    \label{eq:app-expected-grad}
\end{align}

Hence, the expected gradient vanishes when the critic accurately
predicts the expected return,
\(V_{\phi}(s)=\mu_s\).
However, individual sampled returns may still induce nonzero stochastic
updates.

To see this, consider the squared norm of the value-head gradient:
\begin{align}
    \left\|
        \nabla_W \ell
    \right\|_2^2
    &=
    \left\|
        \left(
            V_{\phi}(s)-R
        \right)
        h_\phi(s)
    \right\|_2^2
    \nonumber\\
    &=
    \left(
        V_{\phi}(s)-R
    \right)^2
    \left\|
        h_\phi(s)
    \right\|_2^2.
    \label{eq:app-grad-norm}
\end{align}

Taking the conditional expectation yields
\begin{align}
    \mathbb{E}
    \left[
        \left\|
            \nabla_W \ell
        \right\|_2^2
        \mid s
    \right]
    &=
    \left\|
        h_\phi(s)
    \right\|_2^2
    \,
    \mathbb{E}
    \left[
        \left(
            V_{\phi}(s)-R
        \right)^2
        \mid s
    \right].
    \label{eq:app-second-moment-1}
\end{align}

We can decompose the remaining term by adding and subtracting the
conditional mean \(\mu_s\):
\begin{align}
    V_{\phi}(s)-R
    &=
    \left(
        V_{\phi}(s)-\mu_s
    \right)
    +
    \left(
        \mu_s-R
    \right).
\end{align}

Therefore,
\begin{align}
\mathbb{E}\!\left[(V_\phi(s)-R)^2\mid s\right]
&=(V_\phi(s)-\mu_s)^2
+2(V_\phi(s)-\mu_s)\mathbb{E}[\mu_s-R\mid s]
\nonumber\\
&\quad+\mathbb{E}\!\left[(R-\mu_s)^2\mid s\right].
\label{eq:app-bias-variance-expand}
\end{align}

By definition,
\begin{equation}
    \mathbb{E}
    \left[
        \mu_s-R
        \mid s
    \right]
    =
    \mu_s-\mathbb{E}[R\mid s]
    =
    0,
\end{equation}
while
\begin{equation}
    \mathbb{E}
    \left[
        \left(
            R-\mu_s
        \right)^2
        \mid s
    \right]
    =
    \sigma_s^2.
\end{equation}

Thus,
\begin{equation}
    \mathbb{E}
    \left[
        \left(
            V_{\phi}(s)-R
        \right)^2
        \mid s
    \right]
    =
    \left(
        V_{\phi}(s)-\mu_s
    \right)^2
    +
    \sigma_s^2.
    \label{eq:app-bias-variance}
\end{equation}

Substituting \Cref{eq:app-bias-variance} into
\Cref{eq:app-second-moment-1}, we obtain
\begin{equation}
\boxed{
    \mathbb{E}
    \left[
        \left\|
            \nabla_W \ell
        \right\|_2^2
        \mid s
    \right]
    =
    \left\|
        h_\phi(s)
    \right\|_2^2
    \left[
        \underbrace{
        \left(
            V_{\phi}(s)
            -
            \mathbb{E}[R\mid s]
        \right)^2
        }_{\text{prediction error}}
        +
        \underbrace{
        \operatorname{Var}(R\mid s)
        }_{\text{return variability}}
    \right].
}
\label{eq:app-value-head-second-moment}
\end{equation}

\subsection{Actor-only Overlong Filtering}
\label{app:overlong-filter-analysis}
This analysis supports \Cref{sec:actor-only-filter}.

We analyze a fixed prompt and omit it from the notation.
Let $C$ denote the event that a trajectory is not truncated.
We consider the on-policy gradient without PPO clipping, which characterizes the first-order update around the behavior policy.

\paragraph{Conditional objective under double-sided filtering.}
If the critic discards truncated trajectories, the population minimizer of its squared loss is
\[
V_{\mathrm{filt}}
=
\mathbb{E}_{\pi_\theta}\!\left[
R(\tau)\mid C
\right].
\]
The conditional expected reward is
\[
\mathbb{E}_{\pi_\theta}\!\left[
R(\tau)\mid C
\right]
=
\frac{
\sum_{\tau\in C}\pi_\theta(\tau)R(\tau)
}{
P_{\pi_\theta}(C)
}.
\]
Differentiating the numerator and denominator gives
\begin{equation}
\begin{aligned}
\nabla_\theta\mathbb{E}_{\pi_\theta}[R(\tau)\mid C]
&=\mathbb{E}_{\pi_\theta}\!\left[
R(\tau)\nabla_\theta\log\pi_\theta(\tau)\mid C
\right] \\
&\quad-V_{\mathrm{filt}}\,\nabla_\theta\log P_{\pi_\theta}(C).
\end{aligned}
\end{equation}
The score-function identity also gives
\[
\nabla_\theta\log P_{\pi_\theta}(C)
=
\mathbb{E}_{\pi_\theta}\!\left[
\nabla_\theta\log\pi_\theta(\tau)
\mid C
\right].
\]
Therefore,
\begin{equation}
\mathbb{E}_{\pi_\theta}\!\left[
\bigl(R(\tau)-V_{\mathrm{filt}}\bigr)
\nabla_\theta\log\pi_\theta(\tau)\mid C
\right]
=
\nabla_\theta\mathbb{E}_{\pi_\theta}[R(\tau)\mid C].
\label{eq:conditional-filtered-gradient}
\end{equation}
The conditional critic supplies the normalization correction in this gradient.
Double-sided filtering therefore optimizes expected reward among trajectories that already avoid truncation.

\paragraph{Why the full-distribution critic helps.}
Actor-only filtering instead uses the full-distribution critic
\[
V_{\mathrm{full}}
=
\mathbb{E}_{\pi_\theta}[R(\tau)].
\]
Its filtered actor update decomposes as
\begin{equation}
\begin{aligned}
&\mathbb{E}_{\pi_\theta}\!\left[
\bigl(R(\tau)-V_{\mathrm{full}}\bigr)
\nabla_\theta\log\pi_\theta(\tau)\mid C
\right] \\
&\quad=\nabla_\theta\mathbb{E}_{\pi_\theta}[R(\tau)\mid C]
+\bigl(V_{\mathrm{filt}}-V_{\mathrm{full}}\bigr)
\nabla_\theta\log P_{\pi_\theta}(C).
\end{aligned}
\label{eq:full-critic-filtered-actor}
\end{equation}
When non-truncated trajectories have higher expected reward than the full rollout distribution, the second term directly favors a higher probability of avoiding truncation.
Truncated trajectories do not enter the actor loss, but their returns still affect this signal through the critic baseline.

\paragraph{Selection bias of actor filtering.}
Actor filtering differs from the unfiltered policy gradient because it discards truncated trajectories.
For a fixed baseline $b$, assume the norm of each trajectory-level gradient contribution is bounded by $\Gamma$.
Then
\begin{equation}
\left\|
\begin{aligned}
&\mathbb{E}_{\pi_\theta}\!\left[
(R(\tau)-b)\nabla_\theta\log\pi_\theta(\tau)\mid C
\right] \\
&-\mathbb{E}_{\pi_\theta}\!\left[
(R(\tau)-b)\nabla_\theta\log\pi_\theta(\tau)
\right]
\end{aligned}
\right\|_2
\leq 2\Gamma P_{\pi_\theta}(C^{\mathsf c}).
\end{equation}
The discrepancy therefore vanishes as the probability of truncation approaches zero.

\paragraph{Token-level conditional baselines.}
\label{app:token-filtering}
With Monte Carlo return targets, filtering critic updates changes the squared-loss minimizer at each token state $s_t$:
\[
\underset{v}{\operatorname{arg\,min}}\;
\mathbb{E}\!\left[(R-v)^2\mid s_t,C\right]
=\mathbb{E}[R\mid s_t,C].
\]
For an accurate filtered critic, the expected advantage of token $a_t$ is therefore
\[
\mathbb{E}[R-V_\phi(s_t)\mid s_t,a_t,C]
=\mathbb{E}[R\mid s_t,a_t,C]-\mathbb{E}[R\mid s_t,C].
\]
Each token thus compares its return with the mean among non-truncated continuations from its own state.
These advantages weight the token gradients $\nabla_\theta\log\pi_\theta(a_t\mid s_t)$, which are aggregated to update the shared actor parameters.
Actor-only filtering instead preserves the critic target $\mathbb{E}[R\mid s_t]$, so truncated returns still affect each token's advantage through its baseline.

\subsection{Noise-Normalized Critic Regression}
\label{app:prompt-weighting}
This analysis justifies the prompt-level variance approximation in \Cref{sec:prompt-weighted-critic}.

We show that the prompt-level return variance $\operatorname{Var}(R \mid s)$ upper-bounds the average prefix-level conditional return variance across token states $s_t$ generated from prompt $s$.

\paragraph{Law of Total Variance Across Token Prefixes.}
Fix an initial prompt state $s$ and a generation step $t$.
Let $s_t = (s, y_{<t})$ denote the random token prefix generated by $\pi_\theta$ at step $t$, and let $R$ denote the terminal return of the sampled rollout.
The true state-value function at step $t$ is the conditional expectation $V^{\pi_\theta}(s_t) = \mathbb{E}[R \mid s_t]$.
Conditioning on the initial prompt $s$, the law of total variance decomposes the prompt-level return variance $\sigma_s^2 = \operatorname{Var}(R \mid s)$ into
\begin{equation}
\sigma_s^2
=
\operatorname{Var}(R\mid s)
=
\underbrace{
\mathbb{E}_{s_t\mid s}\!\left[
\operatorname{Var}(R\mid s_t)
\right]
}_{\text{Expected prefix-level return variance}}
+
\underbrace{
\operatorname{Var}_{s_t\mid s}\!\left[
V^{\pi_\theta}(s_t)
\right]
}_{\text{Variance of true state values across prefixes}}.
\label{eq:prompt-variance-decomposition}
\end{equation}
Because $\operatorname{Var}_{s_t \mid s}[V^{\pi_\theta}(s_t)] \ge 0$, the expected prefix-level return variance satisfies the upper bound
\begin{equation}
\mathbb{E}_{s_t\mid s}\!\left[
\operatorname{Var}(R\mid s_t)
\right]
\leq
\sigma_s^2.
\label{eq:residual-variance-bound}
\end{equation}

\paragraph{Bound on the Prefix Value-Gradient Second Moment.}
For the token-level squared value loss $\mathcal{L}_V(\phi) = \frac{1}{2}(V_\phi(s_t) - R)^2$ with stochastic gradient $\mathbf{g}_t = (V_\phi(s_t) - R)\nabla_\phi V_\phi(s_t)$, the conditional expectation of the squared residual at state $s_t$ decomposes as
\begin{equation}
\mathbb{E}\!\left[
\bigl(
V_\phi(s_t)-R
\bigr)^2
\,\middle|\, s_t
\right]
=
\bigl(
V_\phi(s_t)-V^{\pi_\theta}(s_t)
\bigr)^2
+
\operatorname{Var}(R\mid s_t).
\end{equation}
Assuming the critic Jacobian norm is locally bounded by $\|\nabla_\phi V_\phi(s_t)\|_2 \le J$ across prefixes of prompt $s$, taking the expectation over $s_t \mid s$ and applying Inequality~\ref{eq:residual-variance-bound} yields
\begin{equation}
\mathbb{E}\!\left[
\|\mathbf{g}_t\|_2^2
\,\middle|\, s
\right]
\leq
J^2
\mathbb{E}_{s_t\mid s}\!\left[
\bigl(
V_\phi(s_t)-V^{\pi_\theta}(s_t)
\bigr)^2
\right]
+
J^2\sigma_s^2.
\label{eq:prompt-gradient-bound}
\end{equation}
The noise contribution to the prefix-averaged gradient second moment is bounded by $J^2 \sigma_s^2$.
Multiplying the per-prompt critic loss by $w(s)$ scales $\mathbf{g}_t$ by $w(s)$ and changes the return-noise term in the bound in \Cref{eq:prompt-gradient-bound} to $w(s)^2 J^2 \sigma_s^2$.
For positive prompt variance, selecting $w(s) \propto \operatorname{Var}(R \mid s)^{-1/2}$ renders the noise bound $w(s)^2 J^2 \sigma_s^2 \propto J^2$ invariant to the prompt-level return standard deviation.

For discrete rewards with range $\Delta$ and group size $n\geq2$, consider a reference group containing one maximum reward and $n-1$ minimum rewards (or vice versa).
Its empirical variance is $\hat\sigma^2=\Delta^2(1/n)(1-1/n)$, using variance divisor $n$.
We choose the STD floor for groups with identical returns to be approximately half this reference STD:
\[
\frac{\hat\sigma}{2}
=\frac{\Delta\sqrt{n-1}}{2n}
=\frac{\Delta}{2\sqrt n}\sqrt{1-\frac1n}
\approx\frac{\Delta}{2\sqrt n}
=\varepsilon.
\]
Under inverse-STD weighting, their weight relative to the reference group is therefore $\hat\sigma/\varepsilon=2\sqrt{1-1/n}\approx2$.
For the continuous-reward FrontierCS setting, we retain the floor $\varepsilon=0.075$ used in our initial experiments, given the cost of rerunning the experimental suite.

\subsection{Critic Mini-Batch Clipping}
\label{app:minibatch-clipping}
This analysis supports the outlier--noise trade-off in \Cref{sec:minibatch-grad-clip} and the experiment in \Cref{fig:fcs-minibatch-ablation}.

We prove the scaling trends in \Cref{sec:minibatch-grad-clip} using an idealized comparison at fixed critic parameters and regression weights.
Let $z_1,\ldots,z_B$ be independent rollout-gradient contributions with common mean $\mu$ and finite covariance $\Sigma$.
Partition them into $M=B/m$ disjoint mini-batches $\mathcal{I}_k$ of size $m$.
For a fixed clipping threshold $c>0$, define $C_c(g)=g/\max\{1,\|g\|_2/c\}$ and
\begin{equation}
g_k=\frac{1}{m}\sum_{i\in\mathcal{I}_k}z_i,
\qquad
\bar g_c=\frac{1}{M}\sum_{k=1}^{M}C_c(g_k).
\label{eq:app-clipped-gradient-average}
\end{equation}
The normalized average $\bar g_c$ isolates the effect of clipping granularity at a fixed rollout budget.

\paragraph{Single-outlier influence.}
Replace one rollout gradient by an arbitrary vector.
Only its mini-batch gradient $g_j$ changes, to $g_j'$, giving a new average $\bar g_c'$.
Since $\|C_c(g)\|_2\leq c$ for every $g$, the triangle inequality gives
\begin{equation}
\begin{aligned}
\|\bar g_c'-\bar g_c\|_2
&=\frac{1}{M}\|C_c(g_j')-C_c(g_j)\|_2\\
&\leq\frac{\|C_c(g_j')\|_2+\|C_c(g_j)\|_2}{M}
\leq\frac{2c}{M}=\frac{2cm}{B}.
\end{aligned}
\label{eq:app-clipping-outlier-bound}
\end{equation}
At fixed $B$ and $c$, this yields the $O(m)$ outlier-sensitivity bound stated in the main text.
This deterministic bound does not require independence.

\paragraph{Mini-batch gradient noise.}
For the original, uncontaminated gradients, independence gives
\begin{equation}
\begin{aligned}
\operatorname{Cov}(g_k)
&=\frac{1}{m^2}\sum_{i\in\mathcal{I}_k}\operatorname{Cov}(z_i)
=\frac{\Sigma}{m},\\
\mathbb{E}\|g_k-\mu\|_2^2
&=\operatorname{tr}\!\left(\operatorname{Cov}(g_k)\right)
=\frac{\operatorname{tr}(\Sigma)}{m}.
\end{aligned}
\label{eq:app-minibatch-gradient-noise}
\end{equation}
Thus, smaller mini-batches present larger stochastic fluctuations to each clipping operation.
Without clipping, the average over all mini-batches has covariance $\Sigma/B$, independent of $m$.
The $O(1/m)$ trend therefore concerns the noise entering each clipping operation, rather than the variance of the full-batch average.

These results explain the competing effects of mini-batch size in a fixed-parameter model.
Sequential Adam steps recompute gradients and update optimizer state, so \Cref{eq:app-clipping-outlier-bound} does not bound the final parameter change.
The variance identity assumes independent rollout contributions; correlated rollouts or weights estimated jointly from a group may change this scaling.

\section{Additional Experimental Results}
\label{app:additional-results}

\subsection{Best Validation Scores}
\label{app:best-validation-scores}
\Cref{tab:best-validation-scores} reports each method's best unsmoothed validation score from the runs and training horizons in \Cref{fig:main-results}, including critic warmup.
Search-R1 uses the mean over its seven validation datasets at a single checkpoint.
Score gains are computed before rounding, with the second-best method selected separately for each task.

\begin{table}[t!]
\centering
\caption{\textbf{Best validation scores.} Scores use a $0$--$100$ scale; bold and underlining mark the best and second best per task.}
\label{tab:best-validation-scores}
{
\setlength{\tabcolsep}{5pt}
\renewcommand{\arraystretch}{1.08}
\begin{tabularx}{0.92\linewidth}{@{}X*{3}{>{\centering\arraybackslash}p{0.16\linewidth}}@{}}
\toprule
\textbf{Method} & \textbf{FrontierCS} & \textbf{AIME24} & \textbf{Search-R1} \\
\midrule
PPO & 12.90 & \underline{64.06} & 39.48 \\
PPO + actor-only filter & \underline{13.91} & 60.73 & 41.74 \\
HL-Gauss PPO & 10.43 & 63.13 & \underline{42.33} \\
VAPO & 5.58 & 54.90 & 40.36 \\
\midrule
\rowcolor{gblue!6}
\textbf{EasyPPO} & \textbf{14.82} & \textbf{65.52} & \textbf{43.22} \\
\bottomrule
\end{tabularx}

}
\end{table}

\subsection{Critic Diagnostics on AIME and Search-R1}
\label{app:additional-critic-diagnostics}
These results extend \Cref{fig:fcs-critic-diagnostics} in \Cref{sec:experiments}.

\begin{figure}[t]
\centering
\includegraphics[width=\linewidth]{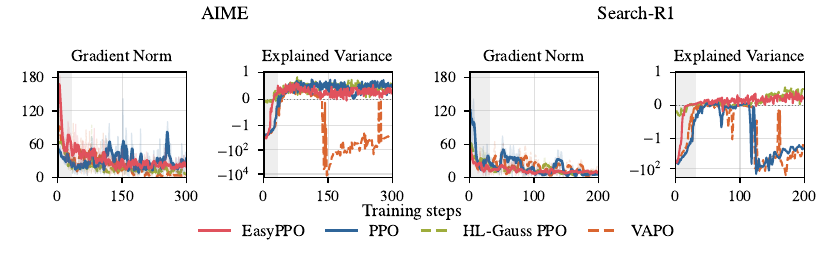}
\caption{\textbf{Critic diagnostics on AIME and Search-R1.}
\textbf{First two panels:} AIME. \textbf{Last two:} Search-R1.
Each pair shows critic gradient norm before clipping and explained variance of value predictions; the latter uses symmetric-log axes.
Gray marks critic warmup.
}
\label{fig:additional-critic-diagnostics}
\end{figure}

\Cref{fig:additional-critic-diagnostics} extends the critic diagnostics to the AIME and Search-R1 runs in \Cref{fig:main-results}.
Gradient norms are measured before clipping and shown with a five-point centered moving average over faint raw values; explained variance is unsmoothed.
On both tasks, EasyPPO avoids the large negative excursions in explained variance seen in VAPO on AIME and in PPO and VAPO on Search-R1.
The behavior remains task-dependent: HL-Gauss retains positive explained variance during its Search-R1 performance decline, so these diagnostics should be interpreted alongside the task scores.

\subsection{Different Critic Initializations on FrontierCS}
\label{app:fcs-repeated-runs}
The runs in \Cref{fig:fcs-seed-comparison} use three different critic initializations per method, with matched core training settings; execution configurations, including GPU count, vary across runs.
Validation is unsmoothed, and every mean and min--max interval uses all three runs over the shared steps $0$--$170$.

\begin{figure}[t]
\centering
\includegraphics[width=0.95\linewidth]{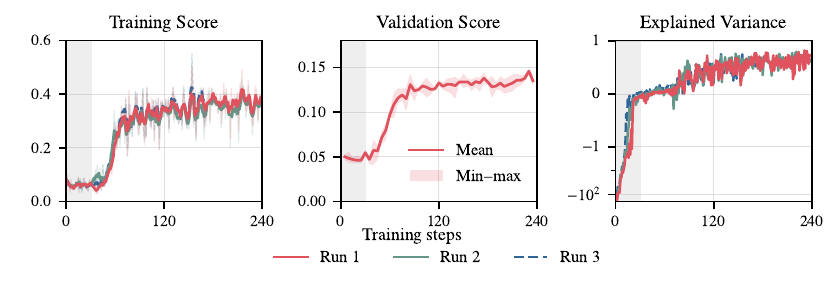}
\caption{\textbf{EasyPPO maintains stable learning across critic initializations.}
FrontierCS. \textbf{Left:} training score. \textbf{Middle:} validation mean and min--max. \textbf{Right:} explained variance.
Gray marks critic warmup.}
\label{fig:fcs-repeated-runs}
\end{figure}

\Cref{fig:fcs-repeated-runs} provides additional EasyPPO diagnostics over the available horizons.
All three runs maintain training gains.
Run~3 ends at step $172$; this figure's validation mean and min--max summarize three runs through step $170$ and two from step $175$ onward, without extrapolation.

\subsection{Offline Return-Variability and Gradient Diagnostics}
\label{app:offline-gradient-diagnostics}
These diagnostics extend \Cref{fig:critic-gradient-scatter} in \Cref{sec:prompt-weighted-critic}.
We analyze $512$ rollouts from each of the checkpoints at steps $30$, $60$, and $90$ in the FrontierCS and AIME training settings, giving $1{,}536$ responses per setting.
Each FrontierCS batch contains $16$ prompts with $32$ responses each; each AIME batch contains $32$ prompts with $16$ responses each.
The latter are training rollouts, not AIME24 validation responses.
The step labels are PPO global training iterations (rollout batches), not individual critic optimizer steps.
The FrontierCS source run shares the model initialization, training data, $512$-rollout batches, group size $32$, $30$-step critic warmup, and $32{,}768$-token response limit of \Cref{fig:frontiercs-overlong-filtering}.
It uses noise-normalized critic learning with four critic mini-batch updates per rollout batch, whereas the actor-only filtering reference in \Cref{fig:frontiercs-overlong-filtering} uses one critic mini-batch without noise normalization.
Using reconstructed response text, we compute the full-critic gradient of terminal-return MSE before clipping, without loss weighting or an optimizer step.
These offline diagnostics isolate the return-regression objective; they do not replay the training-time GAE targets or optimizer updates.

\begin{figure}[t]
\centering
\includegraphics[width=\textwidth]{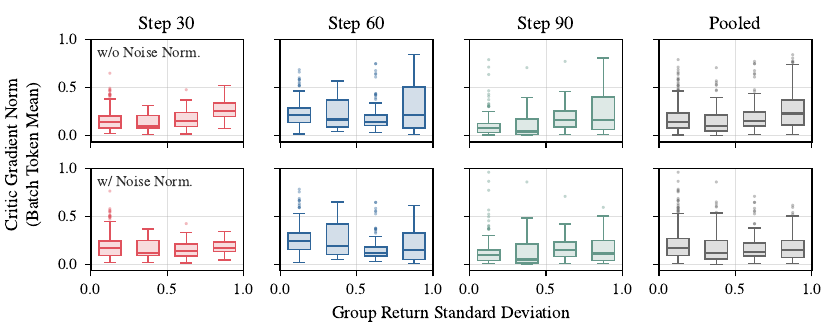}
\captionsetup{font=normalsize}
\caption{\textbf{Noise normalization reduces overall gradient imbalance in the AIME setting.}
Offline diagnostics with $512$ responses per checkpoint ($32$ prompts, $16$ responses each).
\textbf{Columns:} steps $30$, $60$, $90$, and their pooled responses.
\textbf{Top:} without noise normalization. \textbf{Bottom:} with noise normalization.
Boxes summarize per-response critic-gradient norms before clipping, grouped by prompt return standard deviation.}
\label{fig:critic-gradient-boxes-aime}
\end{figure}

For a response with $T_i$ valid tokens, we sum its token-loss gradients and divide by $N=\sum_{j=1}^{512}T_j$, the total valid-token count across all $512$ responses in the same rollout batch.
The plotted quantity is the norm of this response contribution, $\|N^{-1}\sum_{t=1}^{T_i}\nabla_\phi \tfrac12(V_\phi(s_t)-R_i)^2\|_2$.
It is neither a projection onto the batch gradient nor a fraction of its norm: individual gradient vectors can cancel.
Four FrontierCS responses at step $30$ have no valid tokens and contribute zero; we retain them in the plots and group return statistics.

\begin{figure}[t]
\centering
\includegraphics[width=0.94\textwidth]{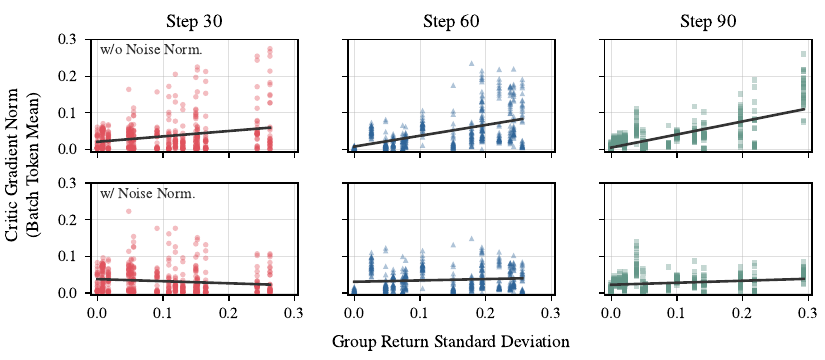}
\captionsetup{font=normalsize}
\caption{\textbf{The FrontierCS gradient-scale trend weakens after noise normalization at each checkpoint.}
\textbf{Columns:} steps $30$, $60$, and $90$.
\textbf{Top/bottom:} the same gradients before/after noise normalization.
Each panel shows $512$ responses; lines are descriptive linear fits.}
\label{fig:critic-gradient-scatter-fcs-steps}
\end{figure}

The horizontal axis is each prompt group's empirical return standard deviation, using population normalization.
For the reweighted panels, we multiply each response gradient by $1/\max\{\hat\sigma(s),\varepsilon\}$, normalized to mean one across prompt groups within that checkpoint.
This retains a comparable overall weight scale for the diagnostic; the batch-token denominator is unchanged.
We use $\varepsilon=0.075$ on FrontierCS and $0.5$ in the AIME setting, whose returns are in $[0,1]$ and $\{-1,+1\}$, respectively.
The appendix scatter plots retain all responses without jitter or trimming; each line is an ordinary least-squares fit with an intercept.
These fits summarize associations, rather than treating responses from the same prompt as independent evidence of causality.
The box plots in \Cref{fig:critic-gradient-scatter} summarize these same FrontierCS responses separately by checkpoint within six fixed return-STD intervals of width $0.05$ spanning $[0,0.30]$.
The intervals are right-closed, with zero included in the first, and contain $17$, $9$, $7$, $7$, $5$, and $3$ checkpoint-specific prompt groups, respectively.
The first three columns correspond to steps $30$, $60$, and $90$; the fourth pools all $1{,}536$ responses. The top and bottom rows show gradients before and after noise normalization, respectively, using identical bins and axis limits.
Boxes show medians and interquartile ranges; whiskers extend to the outermost observations within $1.5$ interquartile ranges of the box, and all points beyond them are shown.

\begin{figure}[t]
\centering
\includegraphics[width=0.94\textwidth]{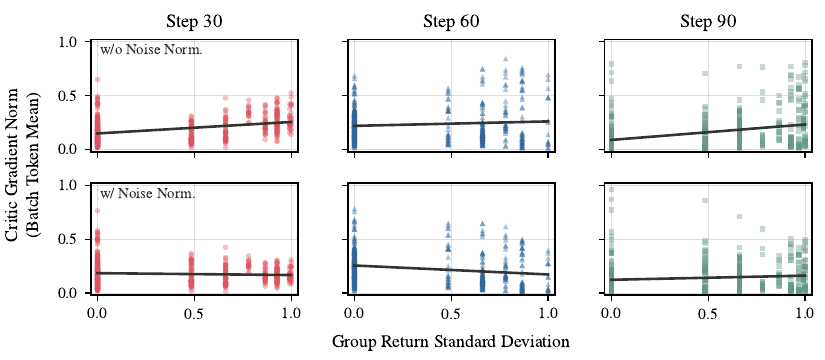}
\caption{\textbf{Response-level gradient variability remains in the AIME training setting.}
\textbf{Columns:} steps $30$, $60$, and $90$.
\textbf{Top/bottom:} the same gradients before/after noise normalization.
Each panel shows $512$ responses; lines are descriptive linear fits.}
\label{fig:critic-gradient-scatter-aime-steps}
\end{figure}

\Cref{fig:critic-gradient-boxes-aime} shows the same comparison in the AIME training setting.
We divide return standard deviations into four equal-width bins over $[0,1]$, containing $46$, $11$, $15$, and $24$ prompt groups across the three checkpoints.
Every bin contains responses at each checkpoint. Finer bins would leave gaps because binary rewards allow only a discrete set of empirical return standard deviations.
Pooling the checkpoints, gradient norms depend less on return variability after normalization, though the effect varies across checkpoints.

\Cref{fig:critic-gradient-scatter-fcs-steps} shows the corresponding per-response scatter plots.
The positive gradient--return-variability trend becomes weaker at each checkpoint, while large individual contributions remain.
\Cref{fig:critic-gradient-scatter-aime-steps} gives the corresponding binary-reward comparison.
Its trends are less uniform: reweighting does not flatten every checkpoint's fit or consistently reduce the largest contribution.
Thus, the diagnostic supports reducing return-scale imbalance, not eliminating all sources of gradient variation.

\begin{figure}[t]
\centering
\includegraphics[width=0.90\linewidth]{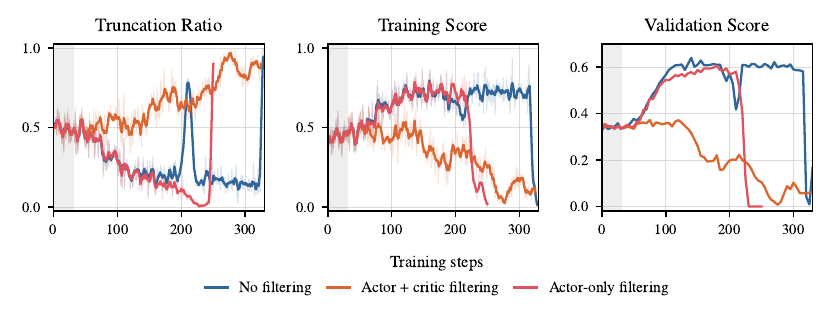}
\caption{\textbf{Joint filtering exhibits increasing truncation and declining performance on AIME.}
\textbf{Left:} truncation ratio. \textbf{Middle:} overall training score. \textbf{Right:} AIME24 validation score.
Gray marks critic warmup. Training statistics use a five-point moving average with faint raw values; validation is unsmoothed.}
\label{fig:additional-filtering}
\end{figure}

\subsection{Overlong Filtering on AIME}
\label{app:additional-filtering}
\Cref{fig:additional-filtering} extends the filtering comparison in \Cref{sec:actor-only-filter} to the AIME setting.
The three runs share Qwen3.5-9B-Base initialization, rollout batches of $512$, group size $16$, an $8192$-token response limit, and a $30$-step critic warmup.
All use one unweighted MSE critic mini-batch with learning rate $2\times10^{-6}$; only the filtering strategy differs in the recorded training settings.
Joint filtering leads to increasing truncation and declining overall training and validation scores.
Actor-only filtering initially tracks the gains of no filtering but later collapses, consistent with filtering alone being insufficient for stability.

\begin{figure}[t]
\centering
\includegraphics[width=0.85\linewidth]{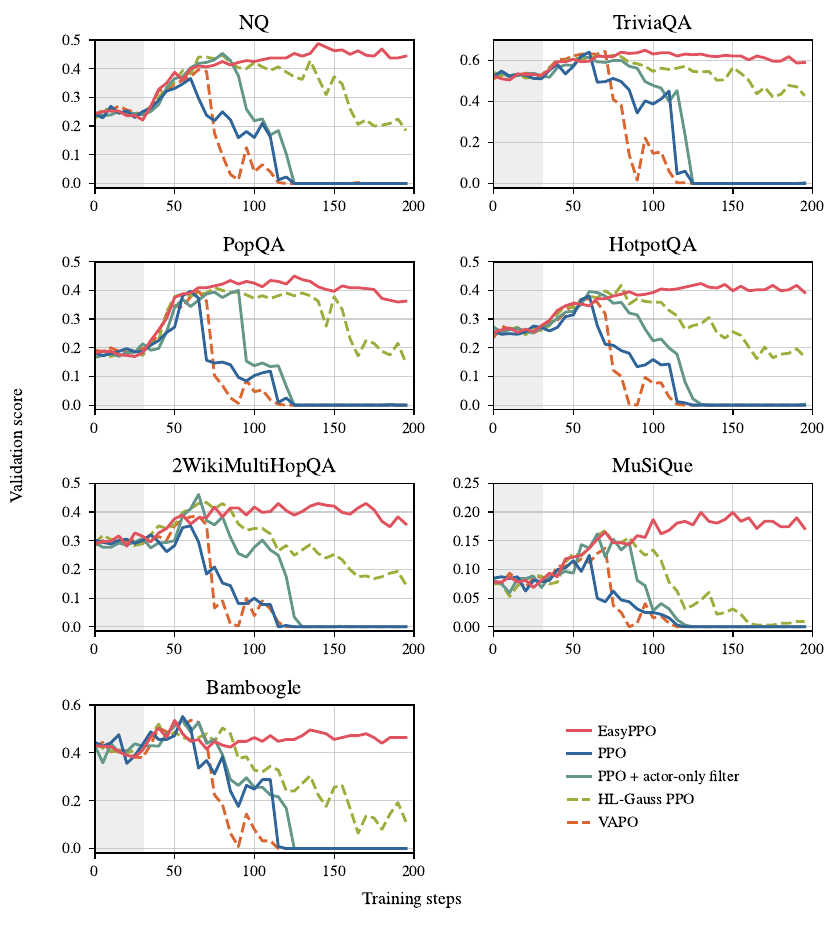}
\caption{\textbf{EasyPPO sustains performance across all seven Search-R1 validation sets.}
Each panel shows one dataset from the aggregate comparison in \Cref{fig:main-results}.
Gray marks the $30$-step critic warmup.}
\label{fig:search-r1-validation-sets}
\end{figure}

\subsection{Search-R1 Results on Individual Validation Sets}
\label{app:search-r1-validation-sets}
\Cref{fig:search-r1-validation-sets} breaks down the seven-dataset Search-R1 average in \Cref{fig:main-results} to examine whether the stability advantage holds across individual datasets.
We use the same five training runs and report unsmoothed validation accuracies under greedy decoding, evaluated every five training steps.

PPO, VAPO, and PPO + actor-only filtering initially improve but later fall to nearly zero on all seven datasets.
HL-Gauss PPO also loses substantial performance, particularly on MuSiQue and Bamboogle.
EasyPPO avoids these collapses and achieves the highest final validation score on every dataset.

\end{document}